%% file: main.tex
\documentclass[10pt,twocolumn,letterpaper]{article}

\usepackage{cvpr}             
\usepackage{booktabs}
\usepackage{tabularx}
\usepackage{amsmath}
\usepackage{multirow}
\usepackage{bbding}
\usepackage{scalerel}
\usepackage{pifont}
\newcommand{\pub}[1]{\color{gray}{\tiny{[{#1}]}}}

\def\thickhline{\noalign{\hrule height.9pt}}
\newcommand{\tabincell}[2]{\begin{tabular}{@{}#1@{}}#2\end{tabular}}
\newcommand{\cmark}{\ding{52}}
\usepackage{bm}
\usepackage{xcolor}
\usepackage[most]{tcolorbox}

\input{listings_style}
\newcommand{\revise}[1]{{\color{black} #1}}


\definecolor{cvprblue}{rgb}{0.21,0.49,0.74}
\usepackage[pagebackref,breaklinks,colorlinks,urlcolor=gray,citecolor=cvprblue]{hyperref}

\def\confName{CVPR}
\def\confYear{2024}

\title{Psychometry: An Omnifit Model for \\ Image Reconstruction from Human Brain Activity}

\author{Ruijie Quan$^{1}$,~ Wenguan Wang$^{1}$\thanks{Corresponding author: \textit{Wenguan Wang}.},~ Zhibo Tian$^{2}$,~ Fan Ma$^{1}$,~ Yi Yang$^{1}$~\\
\small{$^{1}$ ReLER,~ CCAI,~ Zhejiang University, $^{2}$Lanzhou University}\\\small\url{https://github.com/QUANRJ/Psychometry}}

\begin{document}

\maketitle

\input{sec/0_abstract}    
\input{sec/1_intro}

\input{sec/2_relatedwork}

\input{sec/3_methodology}

\input{sec/4_experiments}

\input{sec/5_conclusion}

\clearpage
\setcounter{page}{5}
{
    \small
    \bibliographystyle{ieeenat_fullname}
    \bibliography{main}
}
\input{sec/X_suppl}

\end{document}

%% file: listings_style.tex
\definecolor{codegreen}{rgb}{0,0.6,0}
\definecolor{codegray}{rgb}{0.5,0.5,0.5}
\definecolor{codepurple}{rgb}{0.58,0,0.82}
\definecolor{backcolour}{rgb}{0.95,0.95,0.92}

\lstdefinestyle{mystyle}{
  backgroundcolor=\color{backcolour}, commentstyle=\color{codegreen},
  keywordstyle=\color{magenta},
  numberstyle=\tiny\color{codegray},
  stringstyle=\color{codepurple},
  basicstyle=\ttfamily\footnotesize,
  breakatwhitespace=false,         
  breaklines=true,                 
  captionpos=b,                    
  keepspaces=true,                 
  numbers=left,                    
  numbersep=5pt,                  
  showspaces=false,                
  showstringspaces=false,
  showtabs=false,                  
  tabsize=2
}

%% file: sec/0_abstract.tex
\begin{abstract}
Reconstructing the viewed images from human brain activity
bridges human and computer vision through the Brain-Computer Interface. 
The inherent variability in brain function between individuals leads existing literature to focus on acquiring separate models for each individual using their respective brain signal data, ignoring commonalities between these data. 
In this article, we devise \textit{Psychometry}, an omnifit model for reconstructing images from functional Magnetic Resonance Imaging (fMRI) obtained from different subjects. \textit{Psychometry} incorporates an omni mixture-of-experts (Omni MoE) module where all the experts work together to capture the inter-subject commonalities, while each expert associated with subject-specific parameters copes with the individual differences. 
Moreover, \textit{Psychometry} is equipped with a retrieval-enhanced inference strategy, termed \textit{Ecphory}, which aims to enhance the learned fMRI representation via retrieving from prestored subject-specific memories. 
These designs collectively render \textit{Psychometry} omnifit and efficient, enabling it to capture both inter-subject commonality and individual specificity across subjects.
As a result, the enhanced fMRI representations serve as conditional signals to guide a generation model to reconstruct high-quality and realistic images, establishing \textit{Psychometry} as state-of-the-art in terms of both high-level and low-level metrics. 
\end{abstract}

%% file: sec/1_intro.tex
\vspace{-10pt}
\section{Introduction}
\label{sec:intro}

\begin{figure}[t]
    \centering    
\includegraphics[width=\linewidth]{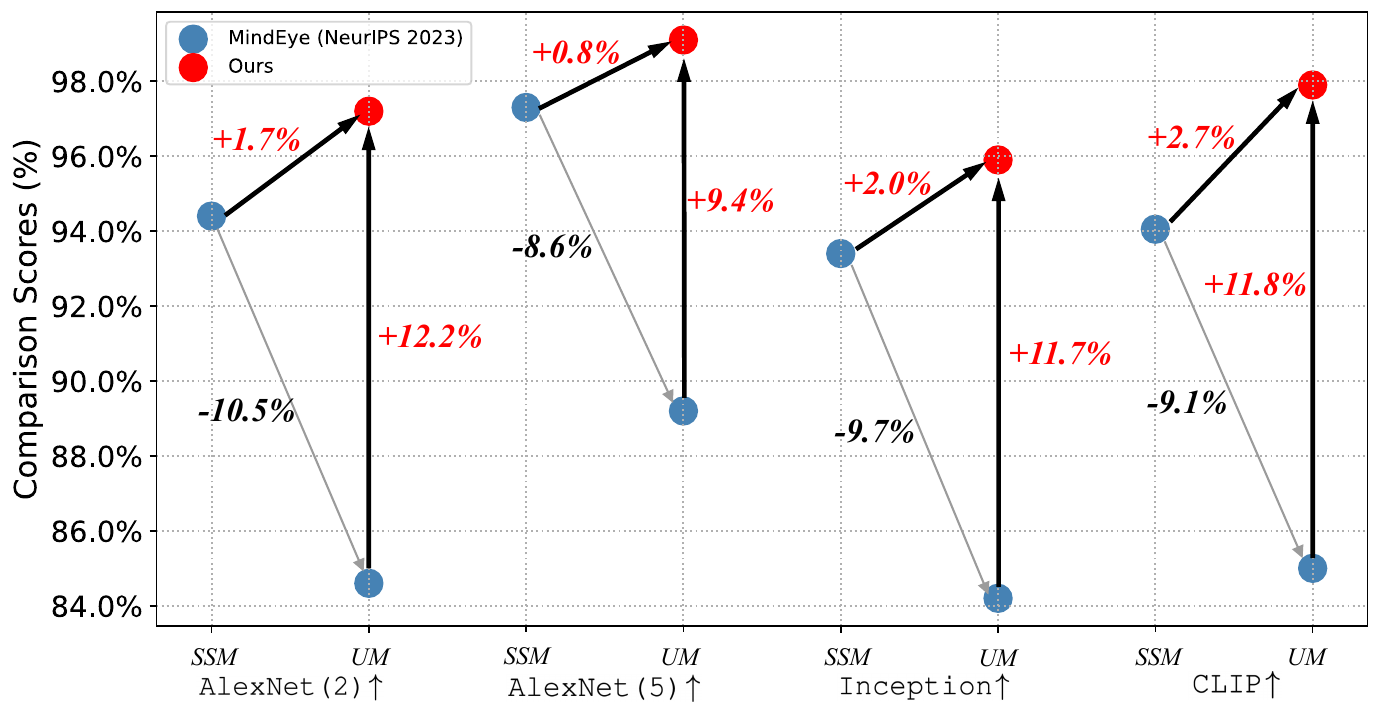}
\includegraphics[width=\linewidth]{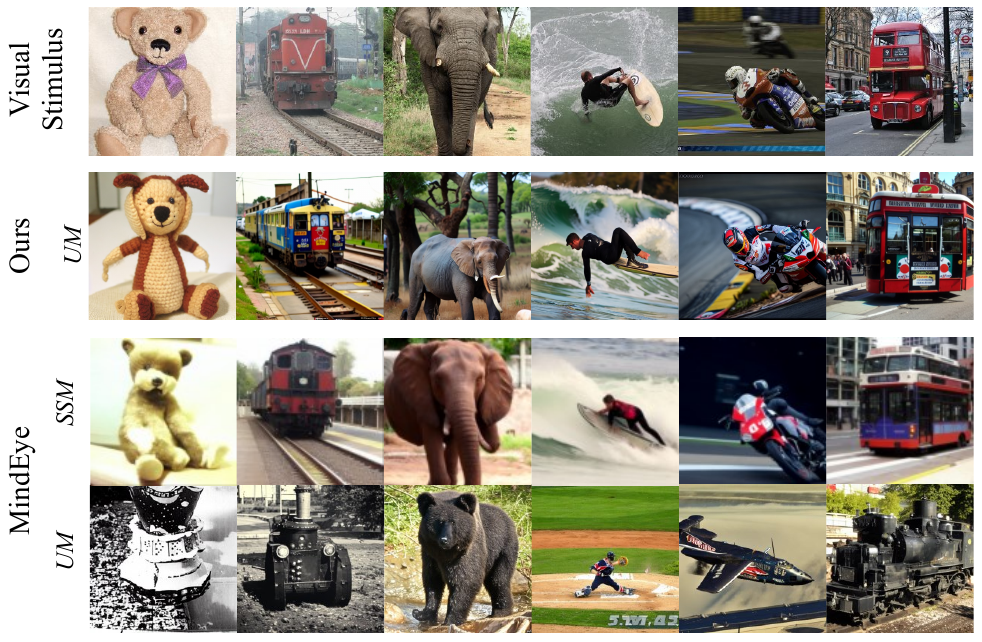}
    \caption{Current fMRI-to-Image methods (\eg, MindEye~\cite{scotti2023reconstructing}) train subject-specific models (\textit{SSM}) on their respective data. They suffer obvious performance degradation when utilizing data from all the subjects to train a unified model (\textit{UM}). Our \textit{Psychometry} enables consistent performance improvements over MindEye by training one omnifit model on the amalgamated fMRI data.
    }\label{fig:fig1_task}
    \vspace{-15pt}
\end{figure}

Understanding the intricacies of brain activity entails extracting meaningful semantics from complex patterns of neural activity~\cite{bullmore2009complex,parthasarathy2017neural,beliy2019voxels,shen2019end}. In the context of visual stimuli, neural responses in the brain are commonly measured by monitoring changes in blood oxygenation using functional Magnetic Resonance Imaging (fMRI)~\cite{kwong1992dynamic}.
Over time, techniques for understanding brain activity based on fMRI have evolved from fMRI classification~\cite{kamitani2005decoding,cox2003functional} to the more challenging fMRI-to-Image reconstruction~\cite{shen2019deep,beliy2019voxels}. 
In neuroscience studies~\cite{fedorenko2010new,nieto2012subject}, individual brains are typically normalized to a template in an attempt to identify common patterns of activation across a group that can be generalized to a given population. 
However, it is evident that brains vary considerably in terms of shape and functional organization among individuals\textemdash normalization cannot fully compensate for these differences~\cite{amunts2015architectonic,gordon2017individual,salvo2021correspondence}. Furthermore, even if anatomical features are perfectly aligned, the same functional region may not occupy the same anatomical region in different participants~\cite{shen2019deep,frost2012measuring}.

The inherent variability in brain functioning across individuals adds complexity to interpreting brain activity.
As a result, \textit{all} existing fMRI-to-Image studies~\cite{lin2022mind,scotti2023reconstructing,gu2022decoding,chen2023seeing,takagi2023high} 
delve into individual subject-specific characteristics by training separate models for each individual using their respective brain signal data.
While these methods undeniably enhance the accuracy and semantic consistency of visual stimulus image reconstruction, they demand the development of individually tailored models for each subject. Not only does this consume substantial computational resources, but the specialized focus on individual differences may also potentially obscure the opportunity to uncover common patterns and similarities among subjects. Consequently, the exploration of broader inter-subject commonality largely remains uncharted territory.
The most straightforward strategy is to amalgamate fMRI data from different subjects for training. Surprisingly, we find that state-of-the-art fMRI-to-Image methods~\cite{scotti2023reconstructing} suffer obvious performance degradation when utilizing data from all the subjects to train a unified model, as illustrated in Figure~\ref{fig:fig1_task}. 
This divergence from the expected benefits of data scaling in improving deep learning model stability and performance~\cite{taylor2018improving} reveals the challenge of building a generalized model for diverse subjects, given their inherent individual differences.

To address this challenge, we propose \textit{Psychometry}, an omnifit model for reconstructing images from fMRI data of various individuals. \textit{Psychometry} can capture both the inter-subject commonalities and the individual variabilities through two essential components. \textbf{First}, drawing inspiration from the powerful concept of Mixture-of-Experts (MoE)~\cite{jacobs1991adaptive,shazeer2017outrageously}, 
\textit{Psychometry} is equipped with an \textit{Omni MoE} module, where all experts participate in the process of fMRI representation learning in order to capture the commonalities from fMRI data among subjects. Moreover, each expert is associated with subject-specific parameters aimed at addressing individual differences. In addition, Omni MoE adopts a \textit{split-then-lump} mechanism with learnable splitting and lumping weights to maintain efficiency. \textbf{Second}, \textit{Psychometry} employs a retrieval-enhanced inference strategy, termed \textit{Ecphory}.
This strategy retrieves the most relevant image or text CLIP~\cite{radford2021learning} embedding from pre-stored training data (referred to as ``memories'') to enhance the learned fMRI representation via a mix-up approach.
The enhanced representations serve as reliable conditional signals to guide a pretrained diffusion model in reconstructing high-quality and realistic images.

\textit{Psychometry} enjoys a few attractive qualities: \textbf{First}, it significantly reduces the model size, training time, and computational resources required. 
This is achieved by the creation of an omnifit model that can handle fMRI data of different subjects, eliminating the need for separately training tailored models on subject-specific data. 
\textbf{Second}, \textit{Omni MoE} along with the \textit{split-then-lump} mechanism enables \textit{Psychometry} to identify the inter-subject commonality and cope with the individual specificity in an efficient way.
\textbf{Third}, with the help of \textit{Ecphory}, \textit{Psychometry} can further improve the fMRI embedding via incorporating the retrieved reliable information from the prestored subject-specific memories, leading to higher-quality image reconstructions from fMRI data.

 In a nutshell, our contributions are three-fold:
 \begin{itemize}
 	\item We propose \textit{Psychometry}, an omnifit model designed to reconstruct images from fMRI data, representing a shift from separately trained models to a more comprehensive and generalized approach.

 	\item 
 	\textit{Psychometry} is integrated with an \textit{Omni MoE} module, enabling all the experts to collectively identify the inter-subject commonalities and individual specificities among fMRI data from diverse subjects, along with a \textit{split-then-lump} manner to ensure efficiency.

 	\item \textit{Psychometry}$_{\!}$ employs$_{\!}$ a$_{\!}$ retrieval-enhanced$_{\!}$ inference strategy, termed \textit{Ecphory}, which accurately retrieves pertinent ``memories'' based on the acquired fMRI representation.
 
 \end{itemize}

%% file: sec/2_relatedwork.tex
\section{Related Work}
\label{sec:relatedwork}

Our work draws on existing literature in image reconstruction from fMRI and mixture-of-experts.  For brevity, only the most relevant works are discussed.

\noindent
\textbf{Image Reconstruction from fMRI.}
Traditional fMRI-to-Image reconstruction methods~\cite{naselaris2009bayesian,kay2008identifying,fujiwara2013modular} rely on fMRI-image paired data and utilize sparse linear regression to predict features from fMRI.
In recent years, researchers have advanced the reconstruction from fMRI techniques by mapping brain signals to the latent space of generative adversarial networks (GANs)~\cite{lin2022mind,mozafari2020reconstructing,ozcelik2022reconstruction}. 
With the release of multimodal vision-language models~\cite{radford2021learning,yang2024doraemongpt,ma2023temporal,liu2023bird,lu2024zero,li2023efficient,suo2024knowledge}, diffusion models~\cite{ho2020denoising,sohl2015deep,song2019generative,song2020score,rombach2022high,shen2023controllable,zhou2024migc}, and large-scale fMRI datasets~\cite{van2013wu,horikawa2017generic,chang2019bold5000,allen2022massive}, image reconstruction from fRMI has reached an unprecedented level of quality~\cite{rakhimberdina2021natural}.
These diffusion model-based methods~\cite{ozcelik2023brain,scotti2023reconstructing} explore mapping fMRI signals to both CLIP text and image embeddings by adopting individual regression models for each subject, subsequently utilizing the pre-trained diffusion model that accommodates multiple inputs for image reconstruction.

Though impressive, these methods primarily
focus on individual subject analysis;  they train specific models for different subjects on their respective data, thus ignoring the commonalities among these data. This highlights the need for a more universal and generalized framework, which is the core motivation behind this work. Furthermore, unlike previous methods that attempt to strictly align fMRI data to CLIP image or text embeddings, we introduce an inference-enhanced strategy named \textit{Ecphory}. It retrieves the image or text CLIP embedding most relevant to the learned fMRI embedding from the pre-stored training data (memories) to enhance the learned fMRI representation as a reliable conditional signal. Moreover, \textit{Ecphory} can effectively explore the individual specificity in the subject-specific memories.

\noindent
\textbf{Mixture of Experts (MoE).}
MoE initially suggests sharing certain experts at the lower levels and combining them through a gating network~\cite{jacobs1991adaptive}. 
Recently, a sparse-MoE framework~\cite{shazeer2017outrageously} was introduced, which routes each input to a subset of activated experts. This leads to a series of studies focusing on routing strategies within Sparse MoEs~\cite{ma2018modeling,riquelme2021scaling}. In particular,~\cite{fan2022m3vit,chen2023mod,ye2023taskexpert,hazimeh2021dselect} introduce task-specific gating networks to choose different experts for processing information from each task. These methods demonstrate success across various applications, \eg, recommendation system~\cite{ma2018modeling}, natural language processing~\cite{fedus2022switch}, and computer vision~\cite{ahmed2016network,riquelme2021scaling}, although the majority of existing works primarily focus on classification tasks~\cite{chen1999improved,gutta2000mixture,enzweiler2011multilevel}.

This work represents the initial exploration of the application of MoE in the field of reconstructing images from fMRI data, with the aim of capturing inter-subject commonality and individual specificity across subjects.
This concept is akin to multi-task learning, which involves utilizing a general model to handle diverse data. However, unlike multi-task learning MoEs that selectively activate experts to address task-specific attributes for different tasks, our focus is on exploring both the inter-subject commonalities and individual specificities present in the diverse individual fMRI inputs.
To achieve this, we introduce an Omni MoE module, where all the experts work together to cooperatively learn the inter-subject commonality. Simultaneously, their associated subject-specific parameters enable different experts to capture the individual specificity.

%% file: sec/3_methodology.tex
\section{Methodology}
\label{sec:method}

\begin{figure*}[t]
    \centering   
\includegraphics[width=0.99\textwidth]{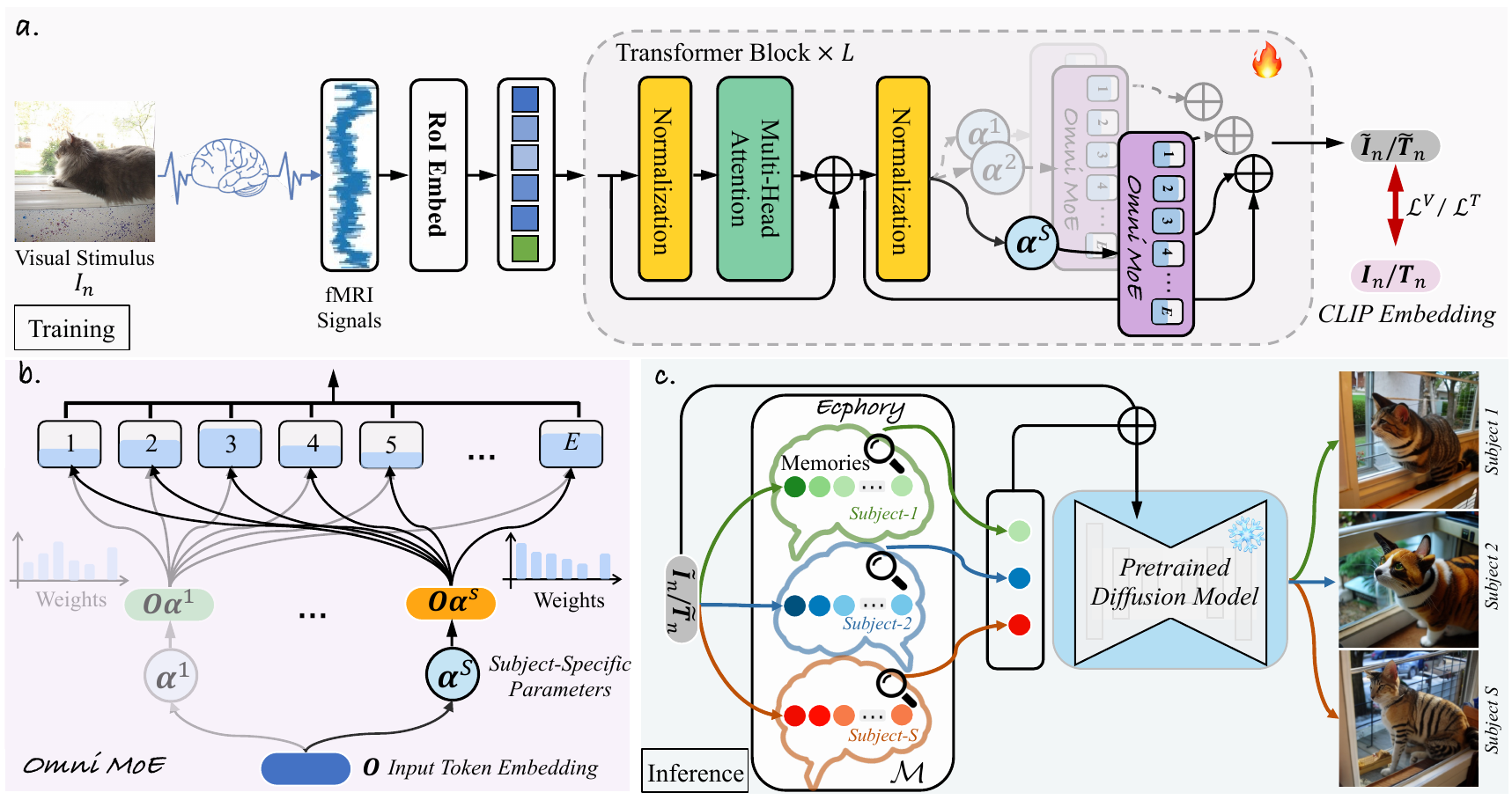}
\vspace{-5pt}
    \caption{(a) Illustration of the \textit{Psychometry} framework (\S\!~\ref{sub_sec:network}). (b) \textit{Omni MoE} engages all experts with subject-specific parameters to work together to capture the inter-subject commonality and individual specificity. The detailed illustration of the ``split-then-lump'' mechanism are presented in Eq.~\ref{eq:2}-Eq.~\ref{eq:5}. (c) 
    \textit{Ecphory} enhances the predicted fMRI embedding by incorporating the retrieved most pertinent ``memories'', serving as more dependable conditional signals to a pre-trained diffusion model. 
    \revise{The reconstruction results for different subjects should align as closely as possible with the visual stimulus, while the inconsistency among the results of different subjects is caused by the individual specificity of each subject's fMRI data.}
    Please refer to \S\!~\ref{sec:method} for more details.} \label{fig:framework}
\vspace{-12pt}
\end{figure*}

\noindent
\textbf{Task Setup and Notations.}
Our target is to reconstruct images from recorded fMRI data as the visual stimulus is presented to a healthy subject. The input fMRI data is usually preprocessed and extracted as a 1D vector of voxels. Formally, let $\boldsymbol{X}_{s,n\!}\!\in\!{\mathbb{R}^d}$ be the input preprocessed fMRI data as an image ${I}_n\!\in\!\mathbb{R}^{H\!\times\!W\!\times\!3}$ was presented to the subject $s\!\in\!\{1,\cdot\cdot\cdot,S\}$, where $d$ is the number of voxels and $n\!\in\!{\{1,\cdot\cdot\cdot,N\}}$. The latent representation of $I_n$ and its corresponding caption text $T_n$ are denoted as $\boldsymbol{I}_n\!\in\!{\mathbb{R}^{v\times{c}}}$ and $\boldsymbol{T}_n\!\in\!{\mathbb{R}^{t\times{c}}}$, respectively, which are obtained by feeding ${I}_n$ and text $T_n$ into CLIP~\cite{radford2021learning}. $v$ and $c$ are the numbers of tokens of the CLIP image and text embeddings while $c$ indicates their dimensions.
Considering the individual variabilities across subjects, existing methods~\cite{liu2023brainclip,lu2023minddiffuser} usually train separate models for each subject using their respective fMRI data, denoted as $\mathtt{f}^{V}_s\!:\!\boldsymbol{X}_{s,n}\!\rightarrow\!{\mathbb{R}^{v\times{c}}}$ and $\mathtt{f}^{T}_s\!:\!\boldsymbol{X}_{s,n}\!\rightarrow\!{\mathbb{R}^{t\times{c}}}$ to predict 
image- or text-aligned fMRI embeddings (${\tilde{\bm{I}}}_n$ and ${\tilde{\bm{T}}}_n$) for each individual subject $s$.
Those predicted features then serve as conditional signals for a diffusion-based model to reconstruct the viewed image ${\tilde{{I}}}_n$.

\noindent
\textbf{Method Overview.}
\textit{Psychometry} is an omnifit model that can explain fMRI data from various subjects. Unlike existing methods necessitate the creation of $S$ models and require training $S$ times for every single modality, \textit{Psychometry} only needs to be trained once using the amalgamated fMRI data, \ie, $\mathtt{p}^{V\!}\!:\!\boldsymbol{X}_{n}\!\rightarrow\!{\mathbb{R}^{v\!\times{c}}\!}$ and $\mathtt{p}^{T}\!:\!\boldsymbol{X}_{n}\!\rightarrow\!{\mathbb{R}^{t\!\times{c}}}$. \textit{Psychometry} involves two core modules: i) an \textit{Omni MoE} layer (\S\!~\ref{sub_sec:FPM}) that exploits inter-subject commonalities and individual specificities;  and ii) a retrieval-enhanced inference strategy (\textit{Ecphory},~\S\!~\ref{sub_sec:RAA}). 
An overview of our complete pipeline can be found in Figure~\ref{fig:framework} and the detailed network architecture is presented in \S\!~\ref{sub_sec:network}.

\subsection{Omni MoE for Learning Inter-Subject Commonality and Individual Specificity}\label{sub_sec:FPM}

\noindent
\textbf{Omni MoE Layer.}
To achieve a full exploration of both the inter-subject commonality and individual specificity from the amalgamated fMRI data of various subjects, our \textit{Psychometry}  is equipped with a  \textit{Omni MoE} layer. Specifically, in \textit{Omni MoE}, there are multiple experts who work in a collaborative manner to capture the commonalities, while each of these experts is assigned a set of subject-specific parameters so as to cope with the individual variabilities. 
Moreover, \textit{Omni MoE} is empowered with a \textit{split-then-lump} mechanism to ease the computational load and prohibit overfitting caused by learning with all experts.
The above designs of our  \textit{Omni MoE} are encapsulated into a network layer and deeply embedded into the Transformer blocks.

Formally, the \textit{Omni MoE} layer contains a group of $E$ experts $f_1,f_2,...,f_E$. 
Given the input sequence tokens $\boldsymbol{O}\!\in\!{\mathbb{R}^{m\times{c}}}$ of a certain Transformer block, where $m$ is the number of tokens and $c$ is their feature dimension, \textit{Omni MoE} works as follows:
\begin{itemize}[leftmargin=*]
	\setlength{\itemsep}{0pt}
	\setlength{\parsep}{-2pt}
	\setlength{\parskip}{-0pt}
	\setlength{\leftmargin}{-10pt}

\item \textit{MOE.} Basically, given the input $\boldsymbol{O}$, \textit{Omni MoE} actively engages all the $E$ experts to generate the output $\boldsymbol{P}$:
\begin{equation}\small
\boldsymbol{P}=\sum\nolimits_{e=1}^{E}f_e(\boldsymbol{O}), 
\label{Eq:MOE}
\end{equation}
where the weights of the $E$ experts $f_1,f_2,...,f_E$ are not shared. Note that each expert needs to process inputs from all the subjects, hence the inter-subject commonalities are explored.

\item \textit{Subject-Specific Parameters.}
In order to further capture individual variants, each expert $f_e$ is associated with a $c$-dimension vector of parameters for each individual subject $s$. The combined parameters are called subject-specific parameters, denoted as $\{\boldsymbol{\alpha}^s\!\in\!\mathbb{R}^{c\times{E}}\}_{1:S}$, \revise{that are only trained with the data of individual subjects, \eg, ${\alpha}^1$ is only optimized by the gradient collected from the fMRI data of subject 1.}
Given $\boldsymbol{O}$ and subject-specific parameters $\{\boldsymbol{\alpha}^s\}_{1:S}$, the subject-specific features are obtained as $\boldsymbol{O}\!\cdot\!\boldsymbol{\alpha}^s\!\in\!\mathbb{R}^{m\times{E}}$. As such, despite letting every expert explore cross-subject patterns from amalgamated fMRI data (Eq.~\ref{Eq:MOE}), these subject-specific parameters enable experts to address the unique aspects of different subjects.

\item \textit{Split-then-Lump: Split.}
Instead of directly processing the original input with a large feature map through every expert, we adopt a \textit{split-then-lump} mechanism in order to maintain computational efficiency. First, the splitting weights are computed $\bm{\omega}$ by applying a \texttt{softmax} function on all the $m$ input tokens of the subject-specific features $\boldsymbol{O}\!\cdot\!\boldsymbol{\alpha}^s$. The splitting weights refer to the specific weights associated with each token, computed as:
\begin{equation}\small
\bm{\omega}^s_{je}=\frac{\text{exp}\big((\boldsymbol{O}\!\cdot\!\boldsymbol{\alpha}^s)_{je}\big)}{\sum\nolimits_{j'\!=\!1}^{m}\text{exp}\big((\boldsymbol{O}\!\cdot\!\boldsymbol{\alpha}^s)_{j'e}\big)}\in\mathbb{R}^{m\times{E}}. \label{eq:2}
\end{equation}
This allows the input $\boldsymbol{O}$ to be compressed into token-wise feature $\bm{\omega}^{s\top}{\boldsymbol{O}}\!\in\!\mathbb{R}^{E\times{c}}$, suggesting all $E$ experts collectively handle $m$ tokens.
Next, we assign the corresponding expert to tackle the splitted feature and compute the output $\boldsymbol{Q}^s$ as a convex combination of all $m$ input tokens.
\begin{equation}\small
\vspace{-3pt}
\boldsymbol{Q}^s=\sum\nolimits_{e=1}^{E}f_e(\bm{\omega}^{s\top}{\boldsymbol{O}})\in\mathbb{R}^{E\times{c}}. \label{eq:3}
\end{equation}

\item \textit{Split-then-Lump: Lump.}
Then, the lumping weights $\mathbf{C}^s$ denote the results of applying a \texttt{softmax} function over the $E$ experts.
The lumping weights suggest the importance of different experts when lumping the features, formulated as:
\begin{equation}\small	
\vspace{-3pt}
\mathbf{C}^s_{je}=\frac{\text{exp}\big((\boldsymbol{O}\!\cdot\!\boldsymbol{\alpha}^s)_{je}\big)}{\sum\nolimits_{e'\!=\!1}^{E}\text{exp}\big((\boldsymbol{O}\!\cdot\!\boldsymbol{\alpha}^s)_{je'}\big)}\in\mathbb{R}^{m\times{E}}.   \label{eq:4}
\vspace{-2pt}
\end{equation}
Finally, the output sequence tokens $\boldsymbol{P}^s$ for subject $s$ is derived as a convex combination from all the $E$ experts, utilizing the computed lumping weights:
\begin{equation}\small
\vspace{-3pt}
\boldsymbol{P}^s=\mathbf{C}^s\boldsymbol{Q}^s\in\mathbb{R}^{m\times{c}}. \label{eq:5}
\vspace{-3pt}
\end{equation}
\end{itemize} 
By utilizing the \textit{split-then-lump} mechanism, \textit{Omni MoE} layer enjoys an efficient approach via separate learning of tokens and dimensions. \revise{Specifically, \textit{split-then-lump} distributes $m$ tokens to $E$ experts (Eq.~\ref{eq:2} \& Eq.~\ref{eq:3}), where $E\!\ll\!{m}$. Then, a comprehensive feature $\boldsymbol{P}^s\!\in\!\mathbb{R}^{m\times{c}\!}$ is derived from the compressed one $\boldsymbol{Q}^s\!\in\!\mathbb{R}^{E\times{c}\!}$ through the lumping operation (Eq.~\ref{eq:4} \& Eq.~\ref{eq:5}). }

\noindent
\textbf{Discussion.}
Existing MoEs used in multi-task learning~\cite{fan2022m3vit,chen2023mod} typically employ a routing network $\mathcal{R}$ to determine task routings via sparsely activating top-$K$ experts~\cite{shazeer2017outrageously} with the largest scores. 
Although these sparse MoEs can offer substantial computational savings, the discrete procedure may introduce biases in activating specific experts based on task-specific attributes, which contradicts our objective of capturing both commonalities and differences in fMRI data from different subjects. 
To address this challenge, \textit{Omni MoE} layer engages all $E$ experts to actively participate in the process of fMRI representation learning, \revise{\ie, each expert $f_e$ in MOE (Eq.~\ref{Eq:MOE}) is required to process the fMRI data from all the subjects to capture inter-subject commonalities.
Concurrently, their associated subject-specific parameters $\{\boldsymbol{\alpha}^s\}_{1:S}$ cope with the individual variabilities by only being trained using the data from individual subjects.}
Note that traditional dense MoE methods also leverage all $E$ experts for learning, but they suffer from intensive computational costs since 
each expert in dense MoE is responsible for processing every input, along with the burden of handling the extensive parameter size in the router. 
Quantitative analyses are later provided in \S\!~\ref{subsec:diaexp}.

\subsection{Ecphory for Test-Time Reconstruction}\label{sub_sec:RAA}
In neurobiological research area~\cite{steinvorth2006ecphory,frankland2019neurobiological}, Ecphory is an automatic memory retrieval process activated when a specific cue interacts with stored information gathered from training data, bringing forth recollections of past events~\cite{tulving1983ecphoric}. 
Drawing inspiration from this concept, we integrate \textit{Psychometry} with a retrieval-enhanced inference strategy named \textit{Ecphory}, where the predicted fMRI embedding serves as the specific cue to interact with the prestored information. 
Specifically, this strategy is tailored to retrieve the most relevant CLIP image or text embedding used as reliable information to enhance the predicted fMRI representation ${\tilde{\bm{I}}}$ and ${\tilde{\bm{T}}}$ rather than directly using them as the conditional signals, \ie, $\boldsymbol{X}_{s,n}\!\rightarrow\!\boldsymbol{I}_n$, $\boldsymbol{X}_{s,n}\!\rightarrow\!\boldsymbol{T}_n$.
This approach is effective as it allows for obtaining a more reliable fMRI representation through a retrieval method, rather than aiming for an ideal alignment to a specific CLIP embedding, which is more challenging. The enhanced conditional signals are then utilized to guide the generation process of a latent diffusion-based model.

\noindent
\textbf{Prestored Subject-Specific Memories.}
Since each subject retains unique memories,  a straightforward approach involves retrieving the relevant embedding from subject-specific memories.
Specifically, we prestore the image CLIP embeddings $\boldsymbol{I}$ and text CLIP embeddings $\boldsymbol{T}$ from different subjects during training as the prestored subject-specific memories, \ie, $\mathcal{M}^{V}\!=\!\{\boldsymbol{I}_n\}_{n=1}^{N}$ and $\mathcal{M}^{T}\!=\!\{\boldsymbol{T}_n\}_{n=1}^{N}$.$_{\!}$ 
Note$_{\!}$ that$_{\!}$ the memories are directly derived from the respective training data and stored before inference.

\noindent
\textbf{\textit{Ecphory} Mechanism.}
With the prestored subject-specific memories, 
the predicted fMRI embeddings ${\tilde{\bm{I}}}_n$ and ${\tilde{\bm{T}}}_n$ act as the specific cues to activate the memory retrieval process. In practice, they are used as queries to retrieve the most relevant CLIP embedding from the subject-specific memories based on their similarities. 
These similarities, denoted as, $sim({\tilde{\bm{I}}}_n, \bm{m}^V), \forall{\bm{m}^V\!\in\!{\mathcal{M}^{V}}}$ and $sim({\tilde{\bm{T}}}_n, \bm{m}^T), \forall{\bm{m}^T\!\in\!{\mathcal{M}^{T}}}$, are computed via a cosine similarity function. In order to utilize the retrieved embedding (denoted as ${\bm{F}}_n$), we employ a mixed-up approach to enrich the fMRI embedding by blending the retrieved embedding, \ie, $\alpha\cdot{\tilde{\bm{I}}_n}+(1-\alpha)\cdot{{\bm{F}}_n}$, where $\alpha$ is a hyperparameter. Consequently, the mixed-up embeddings act as conditional signals to steer the reconstruction process of the pretrained Versatile Diffusion. Given that the diffusion model was initially trained with the CLIP embeddings, our retrieved CLIP embedding could offer more dependable information to the learned fMRI embedding. This process is akin to the ecphory psychology process, where the retrieved ``memory'' aims to evoke recollections of viewed images.

\subsection{Detailed Network Architecture} \label{sub_sec:network}
We adopt the Vision Transformer architecture~\cite{dosovitskiy2020image} as our backbone, in which \textit{Omni MoE} layer is inserted into the transformer block.
As in~\cite{yosinski2014transferable,lecun2015deep}, lower layers in deep neural networks tend to learn more generic information than higher layers, we can reduce computational overhead by applying the \textit{Omni MoE} layer solely to the higher layers. In our experiments, we incorporate the \textit{Omni MoE} layer into the last four out of the twelve transformer blocks.

\noindent
\textbf{Contrastive Learning.}
In practice, \textit{Psychometry} model is trained via treating fMRI as an additional modality, aiming to pull the fMRI embeddings closer to the CLIP space. Given image and text embeddings $\boldsymbol{{I}}_n$ and $\boldsymbol{{T}}_n$ extracted by CLIP, the training objective is to minimize the embedding distances of $({\tilde{\bm{I}}}_n$, $\boldsymbol{I}_n)$ and $({\tilde{\bm{T}}}_n$, $\boldsymbol{T}_n)$. Formally, the employed bidirectional contrastive learning loss is formulated as:
\begin{equation}\small
\begin{aligned}
  \!\!\mathcal{L}^{T\!\!}&\!=\!- \log \frac{\exp \big({\tilde{\bm{T}}}_n^{\top}\boldsymbol{T}_n / {\tau}\big)}{\sum\limits_{j=0}^J \exp \big({\tilde{\bm{T}}}_n^{\top}\boldsymbol{T}_j / \tau\big)}-\log \frac{\exp \big({\tilde{\bm{T}}}_n^{\top}\boldsymbol{T}_n / {\tau}\big)}{\sum\limits_{j=0}^J \exp \big({\tilde{\bm{T}}}_j^{\top}\boldsymbol{T}_n / \tau\big)},  \\ \!\!\mathcal{L}^{V\!\!}&\!=\!- \log \frac{\exp \big({\tilde{\bm{I}}}_n^{\top}\boldsymbol{I}_n / {\tau}\big)}{\sum\limits_{j=0}^J \exp \big({\tilde{\bm{I}}}_n^{\top}\boldsymbol{I}_j / \tau\big)}-\log \frac{\exp \big({\tilde{\bm{I}}}_n^{\top}\boldsymbol{I}_n / {\tau}\big)}{\sum\limits_{j=0}^J \exp \big({\tilde{\bm{I}}}_j^{\top}\boldsymbol{I}_n / \tau\big)},
  \label{eqn:contrastive}
\end{aligned}
\end{equation}
where $\tau$ is a temperature hyperparameter. The sum for each term is over one positive and $J$ negative samples.
The sum over samples of the batch size is omitted for brevity.

%% file: sec/4_experiments.tex
\section{Experiments}
\label{sec:experiments}

\begin{table*}[!bth]
	\centering
			\resizebox{1\textwidth}{!}{
			\setlength\tabcolsep{3pt}
			\renewcommand\arraystretch{1.0}
	\begin{tabular}{|crl||cccc|cccc|}
	\hline \thickhline
	~&~ & & \multicolumn{4}{c|}{\texttt{Low-Level}} & \multicolumn{4}{c|}{\texttt{High-Level}}\\
	\cline{4-11}\cline{4-11}
	\multicolumn{3}{|c||}{\multirow{-2}{*}{Methods}}
	&{\texttt{PixCorr}}\!~$\uparrow$ &\texttt{SSIM}\!~$\uparrow$ &\texttt{AlexNet(2)}\!~$\uparrow$ &\texttt{AlexNet(5)}\!~$\uparrow$  &{\texttt{Inception}}\!~$\uparrow$ &\texttt{CLIP}\!~$\uparrow$ &\texttt{EffNet-B}\!~$\downarrow$ &\texttt{SwAV}\!~$\downarrow$  \\
	\hline
	\hline
	\parbox[t]{2mm}{\multirow{5}{*}{\rotatebox[origin=c]{90}{\textit{SSM}}}} &\multicolumn{1}{|r}{Mind-Reader$_{\!}$~\cite{lin2022mind}}&\!\!\pub{NeurIPS2022}  & $-$ & $-$ & $-$ & $-$  & $78.2\%$ & $-$ & $-$ & $-$ \\
    &\multicolumn{1}{|r}{Mind-Vis$_{\!}$~\cite{chen2023seeing}}&\!\!\pub{CVPR2023}   & $.080$ & $.220$ &  $72.1\%$ & $83.2\%$  & $78.8\%$ & $76.2\%$ & $.854$ & $.491$ \\
    &\multicolumn{1}{|r}{Takagi~\etal$_{\!}$~\cite{takagi2023high}}&\!\!\pub{CVPR2023}   & $-$ & $-$ &  $83.0\%$ & $83.0\%$  & $76.0\%$ & $77.0\%$ & $-$ & $-$ \\
	&\multicolumn{1}{|r}{Gu~\etal$_{\!}$~\cite{gu2022decoding}}&\!\!\pub{MIDL2023}   & $.150$& ${.325}$ & $-$ & $-$  & $-$ & $-$ & $.862$ & $.465$  \\
		&\multicolumn{1}{|r}{MindEye$_{\!}$~\cite{scotti2023reconstructing}}&\!\!\pub{NeurIPS2023}  &$\textbf{.309}$ & $.323$ & ${94.7}\%$ & ${97.8}\%$ & ${93.8}\%$ & ${94.1}\%$ & ${.645}$ & ${.367}$ \\
  \hline
	\hline
	\parbox[t]{2mm}{\multirow{6}{*}{\rotatebox[origin=c]{90}{\textit{UM}}}}& \multicolumn{1}{|r}{Mind-Reader$_{\!}$~\cite{lin2022mind}}&\!\!\pub{NeurIPS2022}  & $-$ & $-$ & $-$ & $-$  & $66.5\%$ & $-$ & $-$ & $-$ \\
    &\multicolumn{1}{|r}{Mind-Vis$_{\!}$~\cite{chen2023seeing}}&\!\!\pub{CVPR2023}   & $.067$ & $.196$ &  $67.7\%$ & $74.2\%$  & $67.9\%$ & $69.3\%$ & $.898$ & $.513$ \\
    &\multicolumn{1}{|r}{Takagi~\etal$_{\!}$~\cite{takagi2023high}}&\!\!\pub{CVPR2023}   & $-$ & $-$ &  $74.0\%$ & $75.1\%$  & $67.3\%$ & $69.0\%$ & $-$ & $-$ \\
	&\multicolumn{1}{|r}{Gu~\etal$_{\!}$~\cite{gu2022decoding}}&\!\!\pub{MIDL2023}   & $.103$& $.264$ & $-$ & $-$  & $-$ & $-$ & $.892$ & $.508$  \\
	&\multicolumn{1}{|r}{MindEye$_{\!}$~\cite{scotti2023reconstructing}}&\!\!\pub{NeurIPS2023}  &$.129$ & $.255$ & $84.2\%$ & $89.2\%$ & $84.1\%$ & $85.0\%$ & $.812$ & $.487$ \\
\cline{2-11}
  &\multicolumn{1}{|c}{\textsc{Psychometry}}&\!\!\footnotesize{(\textbf{\texttt{ours}})} & $.297$  & $\textbf{.340}$ & $\textbf{96.4}\%$ & $\textbf{98.6}\%$ & $\textbf{95.8}\%$ & $\textbf{96.8}$\% & $\textbf{.628}$ & $\textbf{.345}$  \\
\hline
	\end{tabular}}

		\vspace*{-2pt}
		\caption{Quantitative comparison results (\S\!~\ref{subsec_compare_sota}) on NSD~\cite{allen2022massive}~\texttt{test}. \textit{UM} denotes a unified model trained on the amalgamated fMRI data from all subjects, while \textit{SSM} indicates that subject-specific models are trained on subjects' respective data.}
		\label{table:quantitative}
	\vspace*{-2pt}
\end{table*}

\begin{figure*}[t]
    \centering    
        \vspace{-5pt}
\includegraphics[width=\linewidth]{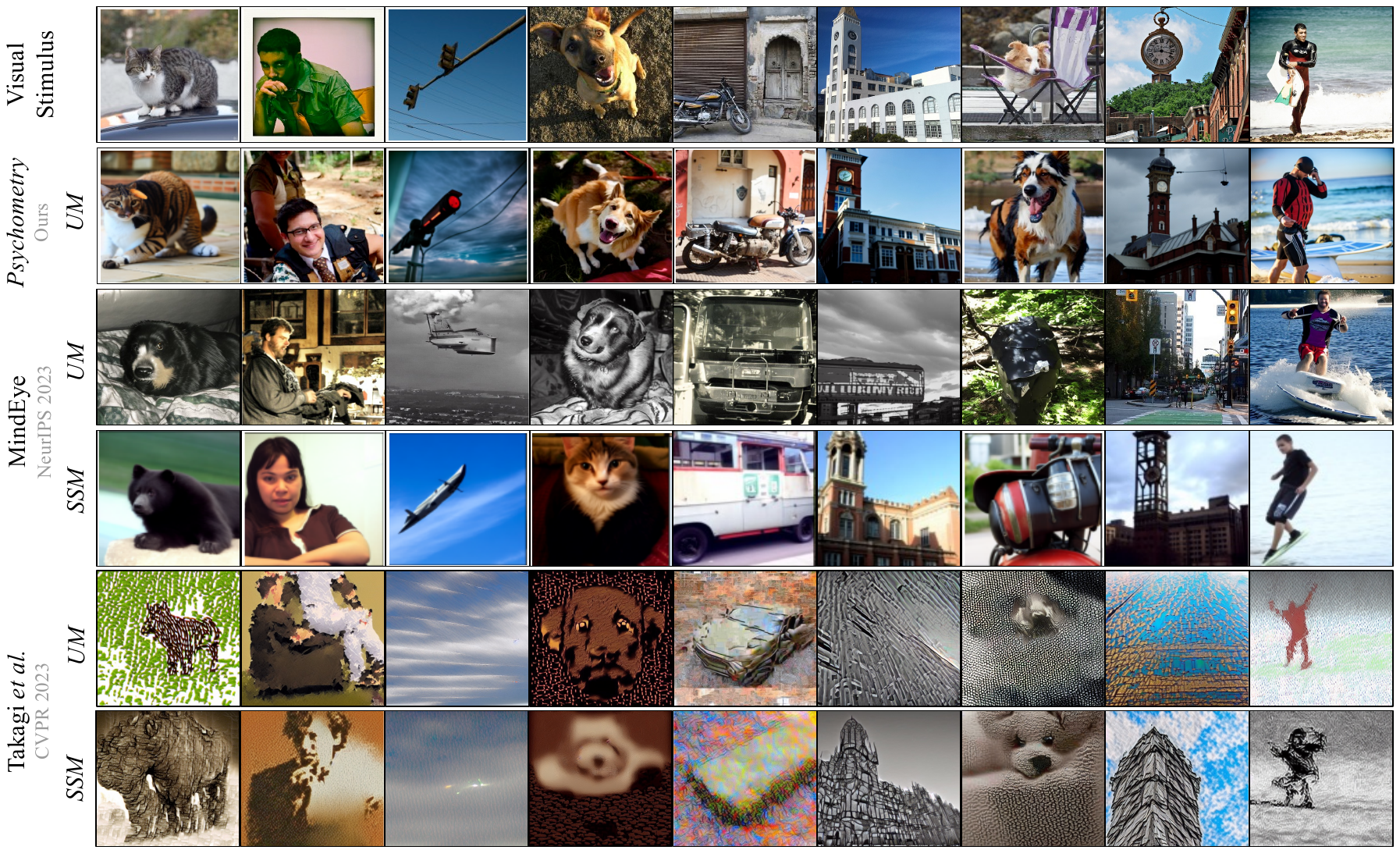}
    \caption{Visual comparison on NSD 
    \texttt{test}. \textit{Psychometry} trains only one unified model (\textit{UM}) for once on the amalgamated fMRI data but generates more accurate reconstructions, even compared to two recent methods~\cite{scotti2023reconstructing,takagi2023high} that train subject-specific models (\textit{SSM}) on their respective data. See \S\!~\ref{subsec_compare_sota} for more detailed discussion.}
    \label{fig:vis_all}
    \vspace{-10pt}
\end{figure*}

\begin{table*}[t]
	\centering
			\resizebox{1\textwidth}{!}{
			\setlength\tabcolsep{3pt}
			\renewcommand\arraystretch{1.0}
	\begin{tabular}{|c|c|c|c||cccc|cccc|}
	\hline \thickhline
	~&~ & & & \multicolumn{4}{c|}{\texttt{Low-Level}} & \multicolumn{4}{c|}{\texttt{High-Level}}\\
	\cline{5-12}\cline{5-12}
	\multirow{-2}{*}{\#} & \multirow{-2}{*}{\tabincell{c}{\textit{Omni}\\\textit{MoE}}} & \multirow{-2}{*}{\tabincell{c}{\textit{Subject-Specific}\\\textit{Parameters}}} & \multirow{-2}{*}{\textit{Ecphory}} 
	&{\texttt{PixCorr}}\!~$\uparrow$ &\texttt{SSIM}\!~$\uparrow$ &\texttt{AlexNet(2)}\!~$\uparrow$ &\texttt{AlexNet(5)}\!~$\uparrow$  &{\texttt{Inception}}\!~$\uparrow$ &\texttt{CLIP}\!~$\uparrow$ &\texttt{EffNet-B}\!~$\downarrow$ &\texttt{SwAV}\!~$\downarrow$  \\
	\hline
	\hline
	 1&&&  & $.163$ & $.238$ &  $74.7\%$ & $85.2\%$  & $80.8\%$ & $81.6\%$ & $.856$ & $.471$ \\
	 2& \cmark&& \cmark & $.237$ & $.287$ & $89.5\%$ & $90.9\%$  & $86.7\%$ & $87.2\%$ & $.794$ & $.423$ \\

    3&\cmark &\cmark &  & $.279$ & $.317$ &  $94.9\%$ & $97.0\%$  & $93.7\%$ & $94.8\%$ & $.647$ & $.365$ \\
		4&\cmark &\cmark &\cmark &  $\textbf{.297}$  & $\textbf{.340}$ & $\textbf{96.4}\%$ & $\textbf{98.6}\%$ & $\textbf{95.8}\%$ & $\textbf{96.8}$\% & $\textbf{.628}$ & $\textbf{.345}$ \\

\hline
	\end{tabular}}

		\vspace*{-2pt}
		\caption{Ablation study on NSD~\cite{allen2022massive}~\texttt{test}. See related analysis in \S\!~\ref{subsec:diaexp}.}
		\label{table:ablation}
	\vspace*{-12pt}
\end{table*}

\subsection{Experimental Setup}

\noindent
\textbf{Datasets.}
Natural Scenes Dataset (NSD)~\cite{allen2022massive} comprises fMRI data collected from 8 participants who viewed a total of 73,000 RGB images. 
This dataset has been widely utilized~\cite{lin2022mind,chen2023seeing,takagi2023high,gu2022decoding,scotti2023reconstructing} to reconstruct perceived images from fMRI.
Following the standard setting, we use the data from subjects $1$, $2$, $5$, and $7$, who completed all the designed trials, \ie, viewed 10,000 natural scene images and repeated 3 times.
We train and evaluate our method using the exact same data split as previous studies.
Specifically, the \texttt{train} set for each subject contains 8,859 image stimuli and 24,980 fMRI trials. The \texttt{test} set includes 982 image stimuli and 2,770 fMRI trials.
All images and captions are sourced from MS-COCO database~\cite{lin2014microsoft}. Different from previous methods which separately train the network for each subject, our proposed method jointly learns the training set for all subjects.

\noindent
\textbf{Evaluation Metrics.}
Both qualitative and quantitative evaluations are conducted in our experiments. For qualitative evaluation, we visually compare our reconstructed images with the ground truth images and the results of the state-of-the-art methods in Figure~\ref{fig:vis_all}. For quantitative evaluation, we employ eight metrics for high-level and low-level evaluation following established research~\cite{scotti2023reconstructing,gu2022decoding,takagi2023high}. 
Specifically, high-level metrics consist of the latent distance of EffNet-B~\cite{tan2019efficientnet} and SwAV~\cite{caron2020unsupervised}, which quantifies the similarity between artificial neural networks and the brain's mechanisms for core object recognition.
Low-level metrics include the classical Structural Similarity (SSIM) and pixel-wise correlation (PixCorr).

\noindent
\textbf{Reproducibility.}
Our model is implemented in PyTorch and trained on one NVIDIA RTX A6000 GPU with a 48GB memory. Testing is conducted on the same machine.

\subsection{Comparison to State-of-the-Arts}\label{subsec_compare_sota}
\noindent
\textbf{Quantitative Results.}
We compare \textit{Psychometry} with five state-of-the-art methods, namely Mind-Reader~\cite{lin2022mind}, Mind-Vis~\cite{chen2023seeing}, Takagi~\etal~\cite{takagi2023high}, Gu~\etal~\cite{gu2022decoding}, and MindEye~\cite{scotti2023reconstructing}. 
As shown in Table~\ref{table:quantitative}, \textit{Psychometry} trained on the amalgamated data from all subjects demonstrates competitive results compared to all other baselines. 
In particular, we observe that existing methods show a noticeable performance decrease
when their models are trained on the amalgamated data from all subjects. For instance, MindEye~\cite{scotti2023reconstructing} shows a significant decrease $\textbf{58.3}\%/\textbf{21.1}\%/\textbf{10.5}\%/\textbf{8.6}\%$ in all low-level metrics, suggesting that these methods severely suffer from the individual specificities across subjects. 
However, our method earns $\textbf{12.2}\%$, $\textbf{9.4}\%$, $\textbf{11.7}\%$, and $\textbf{11.8}\%$ performance gains over MindEye~\cite{scotti2023reconstructing}, which is current state-of-the-art, in terms of \texttt{AlexNet(2)}, \texttt{AlexNet(5)}, \texttt{Inception}, and \texttt{CLIP} respectively. Compared to Takagi~\cite{takagi2023high}, our method  significantly lifts the scores by $\textbf{28.5}\%$ and $\textbf{27.8}\%$ on two high-level metrics.
Note that the results of ``\textit{SSM}'' in Table~\ref{table:quantitative} are averaged from four subject-specific models, each of which is trained on the subject's respective data.
As demonstrated, \textit{Psychometry} 
can still provide notable performance gains when compared to these methods and sets a new state-of-the-art. 
For instance, \textit{Psychometry} promotes MindEye~\cite{scotti2023reconstructing} by $\textbf{2.0}\%/\textbf{2.7}\%/\textbf{2.6}\%/\textbf{6.0}\%$ and Mind-Vis~\cite{chen2023seeing} by $\textbf{17.0}\%/\textbf{20.6}\%/\textbf{26.5}\%/\textbf{29.7}\%$ over the four high-level metrics. These improvements are particularly impressive considering that our method only has to train a single model once on the amalgamated data.

\noindent
\textbf{Qualitative Results.}
As depicted in Figure~\ref{fig:vis_all}, the qualitative results are consistent with the numerical findings,
demonstrating that our approach produces superior quality and more realistic reconstructions compared to the other methods.
In particular, current state-of-the-art, \ie, MindEye~\cite{scotti2023reconstructing}, suffers from a noticeable performance decrease when its models are trained on the amalgamated data from all subjects. This decrease in performance is evident in the reconstructed images, \eg, when it reconstructs a truck while the visual stimulus was a motorbike.
In contrast, the reconstructed images generated by \textit{Psychometry} maintain a high level of consistency with the visual stimuli in terms of semantics, appearance, and structure. This indicates that \textit{Psychometry} can effectively capture inter-subject commonality and individual specificity across subjects, resulting in high-quality image reconstructions from fMRI data.

\subsection{Diagnostic Experiment}\label{subsec:diaexp}
To thoroughly demonstrate how each component in \textit{Psychometry} contributes to the performance, a series of ablation experiments are conducted on NSD \texttt{test} set. All variants are based on ViT~\cite{dosovitskiy2020image} backbone (`\#1' in Table~\ref{table:ablation}).

\begin{figure}[t]
    \centering    
\includegraphics[width=\linewidth]{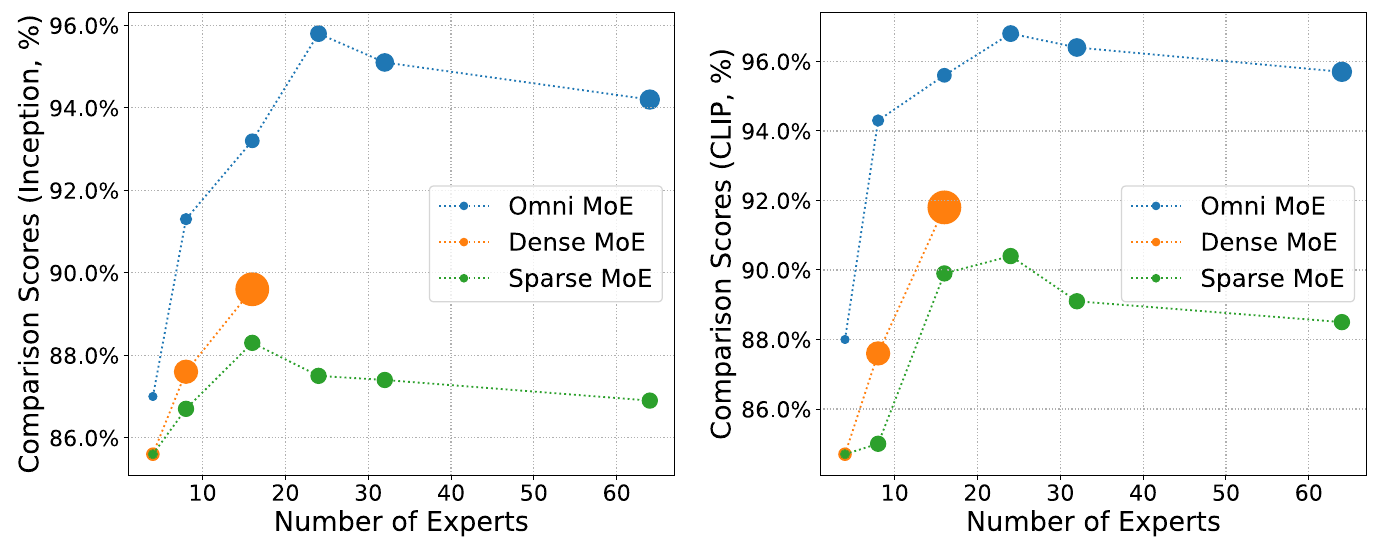}
    \caption{The comparison scores (\texttt{Inception} and \texttt{CLIP}) and the model parameters vary as the number of experts increases. The size of the marker depends on the model size. See \S\!~\ref{subsec:diaexp} for details.}\label{fig:num_exp}
        \vspace{-12pt}
\end{figure}

\noindent
\textbf{Omni MoE Layer.}
We first investigate the effectiveness of the Omni MoE layer which consists of subject-specific parameters and a \textit{split-then-lump} mechanism. As shown in Table~\ref{table:ablation}, the Omni MoE layer bings noticeable performance boost (\eg, $0.163/0.238/74.7\%/85.2\%\!\rightarrow\!{0.279/0.317/94.9\%/97.0\%}$ on all low-level metrics). This suggests that a single shared MLP layer in the baseline backbone is far from enough to capture the inter-subject commonality and tackle the individual specificity, and proves the effectiveness of our Omni MoE layer. In addition, we derive two more variants that replace the omni MoE layer with a sparse MoE (\ie, top$K$~\cite{shazeer2017outrageously}) and a classical dense MoE. The comparison results in Figure~\ref{fig:num_exp} suggest that, although these two variants also boost the performance over the baseline (`\#1' in Table~\ref{table:ablation}), Omni MoE layer outperforms them obviously.

\noindent
\textbf{Subject-Specific Parameters.}
Table~\ref{table:ablation} also investigates the impact of the subject-specific parameters in the Omni MoE layer. When these parameters are not used (labeled as `\#2'), each Omni MoE layer adopts shared parameters across subjects, without considering how to tackle the individual differences across subjects. Doing so leads to worse performance, \ie, $96.8\%\!\!\rightarrow\!\!87.2\%$ over the \texttt{CLIP} scores.

\noindent
\textbf{Computational Efficiency.} 
We further investigate the efficiency gains facilitated by our \textit{split-then-lump} mechanism (Eq.~\ref{eq:2} - Eq.~\ref{eq:5}). As depicted in Figure~\ref{fig:num_exp}, the \textit{Omni MoE} exhibits cost-effective computational overhead, despite involving all its experts in the fMRI representation learning process. This results in a substantial reduction in model size when compared to the variant, dense MoE (Eq.~\ref{Eq:MOE}), even when comprising the same number of experts. Furthermore, \textit{Psychometry} equipped with \textit{Omni MoE} demonstrates comparable parameters while achieving superior performance, even when compared to the variant using sparse MoE.

\noindent
\textbf{Ecphory Inference Strategy.} We then proceeded to assess the effectiveness of our Ecphory inference strategy. The corresponding results are summarized in Table~\ref{table:ablation}. In the absence of this strategy (labeled as `\#3'), 
we directly use the predicted embeddings, \ie, ${\tilde{\bm{I}}}_n$ and ${\tilde{\bm{T}}}_n$, as the conditional signals for the pretrained diffusion model. Consequently, the performance drops $\textbf{0.018}/\textbf{0.023}/\textbf{1.5}\%/\textbf{1.6}\%$ and $\textbf{2.1}\%/\textbf{2.0}\%/\textbf{0.019}/\textbf{0.02}$ across all low-level and high-level metrics, respectively. This evidences that leveraging such a retrieval-enhanced strategy during inference leads to more reliable condition signals and supports our motivation that directly mapping fMRI embeddings to the CLIP image or text embeddings falls short.

\begin{figure}[]
    \centering  
        \vspace{-5pt}
\includegraphics[width=\linewidth]{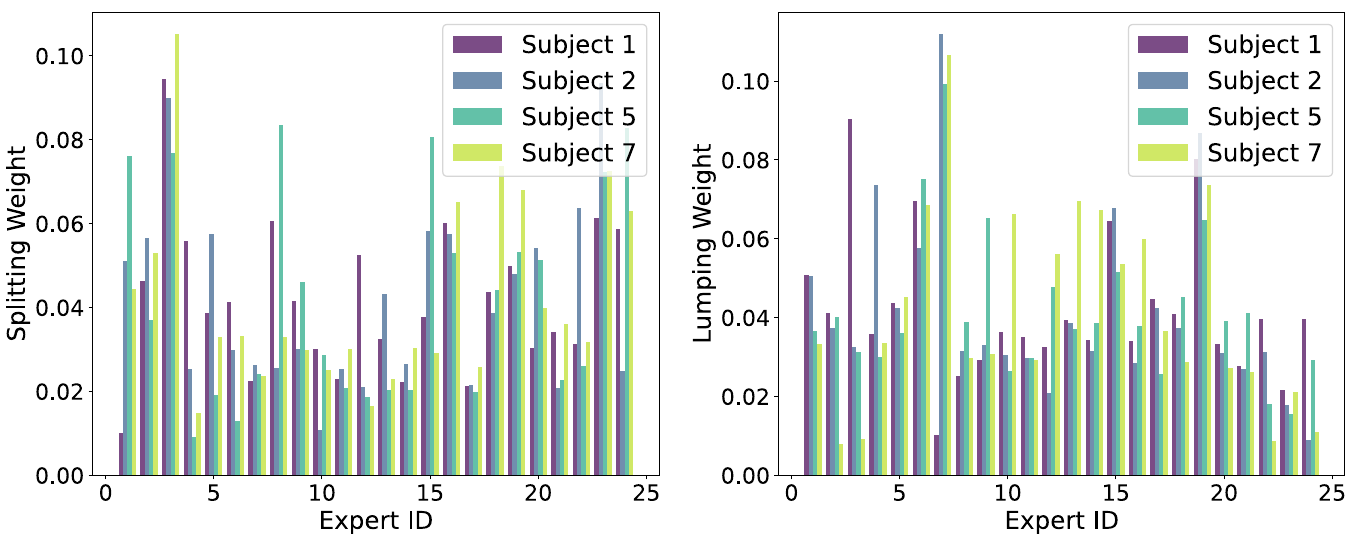}
    \vspace{-18pt}
    \caption{Splitting weights (Eq.~\ref{eq:2}) and lumping weights (Eq.~\ref{eq:4}) across experts for all four subjects. See related analysis in \S\!~\ref{subsec:diaexp}.} \label{fig:weights1}
    \vspace{-10pt}
\end{figure}

\noindent
\textbf{Splitting and Lumping Weights.}
We visualize the splitting weights and lumping weights by summing them across the token dimension, presented in Figure~\ref{fig:weights1}. We observe that some experts in our \textit{Omni MoE} layer have high weights for all subjects, \eg, 3rd and 16th expert in splitting weights, while others vary significantly, providing valuable insights into the model's behavior. This suggests that certain experts are adept at capturing common patterns across all subjects, while others excel at capturing subject-specific nuances. This aligns with the design of our model, where subject-specific parameters enable experts to focus on individual specificity, while the collaborative nature of the Omni MoE layer facilitates the capture of inter-subject commonalities. This balance between commonality and specificity is crucial for the model to effectively learn and generalize from the fMRI data across different subjects.

\noindent
\textbf{Number of Experts.}
As the number of experts increases, the computational cost of the model also rises. We conduct experiments by increasing the number of experts in all three variants and training these models for the same duration to determine the best-performing model. As depicted in Figure~\ref{fig:num_exp}, we discontinue the use of $E\!>\!16$ for Dense MoE (Eq.~\ref{Eq:MOE}) due to memory constraints exceeding the computational limits of our hardware. Sparse MoE does not yield performance gains with the increased number of experts. On the other hand, Omni MoE achieves its peak performance when $E\!=\!24$. However, increasing $E$ above 24 provides marginal or even negative gain. This may be because too many experts would find some insignificant patterns that are trivial or harmful to decision-making. Therefore, we use $E\!=\!24$ in all other experiments.

\begin{figure}[]
    \centering  
    \vspace{-5pt}
\includegraphics[width=\linewidth]{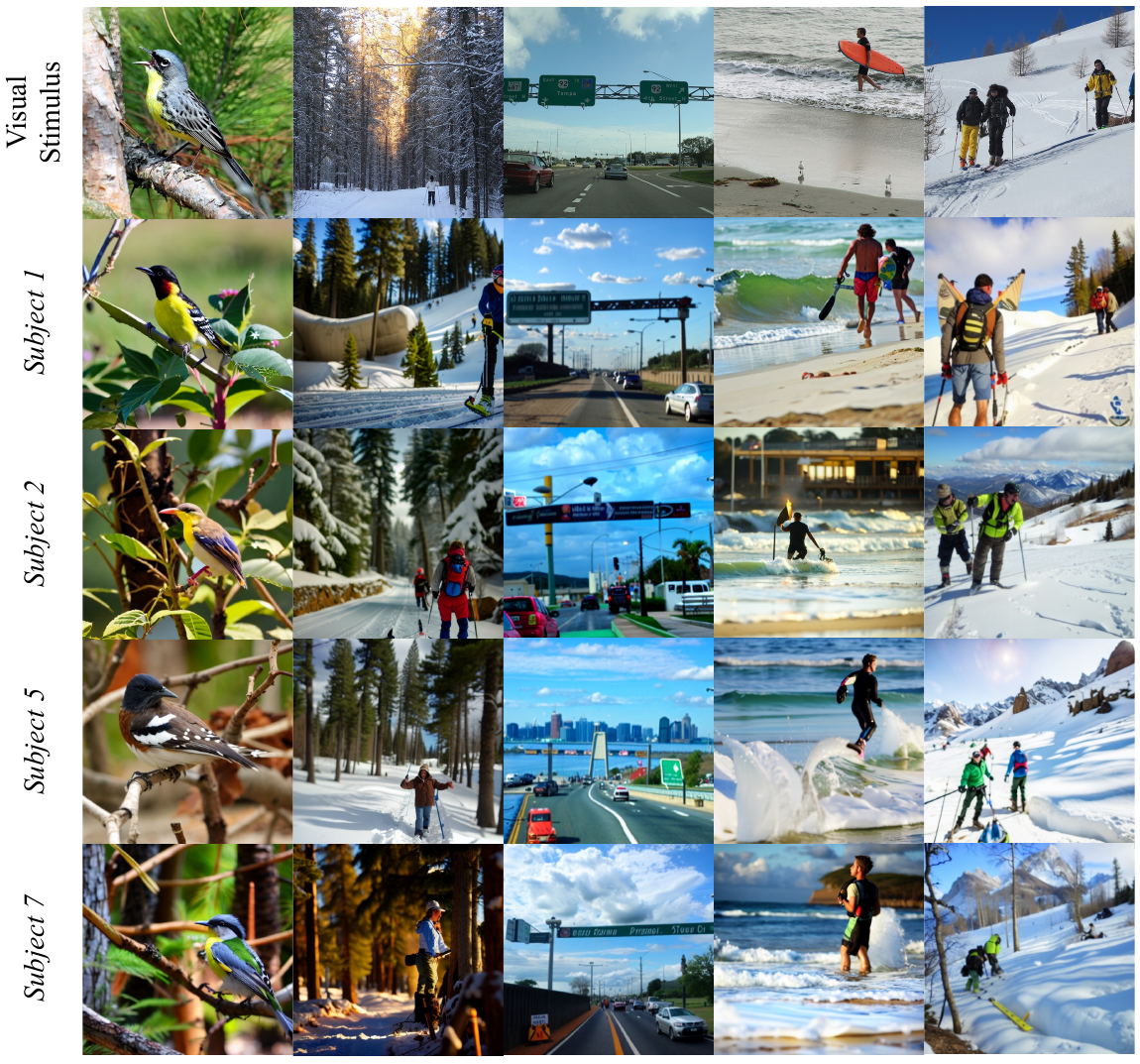}
    \vspace{-18pt}
    \caption{
    Reconstruction results of \textit{Psychometry} for different subjects with the same visual stimuli. See related analysis in \S\!~\ref{subsec:diaexp}.} \label{fig:subjects}
    \vspace{-12pt}
\end{figure}

\noindent
\textbf{Inter-Subject Commonality and Individual Specificity.}
\revise{Figure~\ref{fig:subjects} 
reveals the semantic coherence and visual discrepancies among the reconstruction results of different subjects when exposed to the same visual stimuli. This consistency underscores the proficiency of \textit{Psychometry} in capturing shared patterns across subjects,
while its use of a single model to generate subject-specific outcomes further validates its effectiveness in fMRI-based image reconstruction.
However, the inconsistencies in the results accentuate the individual specificity of each subject's fMRI data.}

%% file: sec/5_conclusion.tex
\section{Conclusion and Discussion}
\label{sec:conclusion}

In this paper, we introduce \textit{Psychometry}, an omnifit model for fMRI representation learning
which marks a significant departure from previous separate training approaches. 
By leveraging the powerful concept of MoE in an efficient Omni MoE and introducing \textit{Ecphory}, a retrieval-enhanced inference strategy, \textit{Psychometry} can efficiently and effectively capture inter-subject commonality and individual specificity across subjects, resulting in high-quality and realistic image reconstructions from fMRI data.
Moving forward, the development of \textit{Psychometry} presents new challenges, particularly in the area of fMRI data privacy protection when amalgamating fMRI data from various subjects for training. Given the rapid advancements in related techniques, we anticipate a surge of innovation towards addressing this promising direction in the field of image reconstruction from human brain activity.

%% file: sec/X_suppl.tex
\clearpage
\setcounter{page}{1}
\maketitlesupplementary
\appendix

\renewcommand{\lstlistingname}{Algorithm}
\crefname{listing}{algorithm}{algorithms}
\Crefname{listing}{Algorithm}{Algorithms}
This supplementary material provides more details of our implementations, the NSD dataset used, fMRI preprocessing techniques, the Omni MoE algorithm, and the Ecphory algorithm, as well as additional reconstruction results, quantitative results, broader impacts, and limitations. These topics are organized as follows:
\begin{itemize}
	\item \S\ref{sec:datasets}: Dataset and fMRI Data Preprocessing Details
	\item \S\ref{sec:reconstruction}: More Reconstruction Results
    \item \S\ref{sec:implementation}: More Implementation Details
	\item \S\ref{sec:broaderimpact}: Social Impacts
	\item \S\ref{sec:limitation}: Limitations

\end{itemize}

\section{Dataset and fMRI Preprocessing Details}\label{sec:datasets}
The Natural Scenes Dataset\footnote{\url{http://naturalscenesdataset.org/}} (NSD)~\cite{allen2022massive} is currently the largest publicly available fMRI dataset. It contains densely sampled fMRI data from 8 participants (2 male and 6 female, age 19-32 years) viewing a total of 73,000 RGB images. Each participant saw a total of 10,000 unique images (repeated 3 times each) across 20 to 40 7T MRI sessions (whole-brain gradient-echo EPI, 1.8-mm iso-voxel and 1.6s TR). Each session consisted of 12 runs of 5 minutes each, where each image was seen for 3 s, with a 1-s blank interval between two successive image presentations. Among the 8 participants, only 4 (namely subject 1, subject 2, subject 5 and subject 7) completed all sessions. The images utilized in the NSD dataset were sourced from the Microsoft Common Objects in Context (COCO) database~\cite{lin2014microsoft}, square-cropped, and displayed at a size of $8.4^\circ\!\times\!{8.4^\circ}$. A set of 982 among them was shared across all subjects while the remaining images for each individual were mutually exclusive across subjects. 
Existing fMRI-to-image reconstruction studies~\cite{scotti2023reconstructing,gu2022decoding,takagi2023high} that used NSD typically follow the same procedure: training subject-specific models for the four participants who finished all scanning sessions and employing a \texttt{test} set that corresponds to the common 982 images shown to each participant.
However, in our experiments, \textit{Psychometry} trains a single model on the amalgamated data for the four subjects. 
Specifically, the \texttt{train} set contains 35,436 images and 99,920 fMRI trials, and the \texttt{test} set consists of 982 images and 11,080 fMRI trials.

Pre-processing of the fMRI data involved performing temporal interpolation to correct slice time differences and spatial interpolation to account for head motion artifacts. Subsequently, a general linear model was employed to estimate single-trial beta weights. The NSD dataset also encompasses cortical surface reconstructions generated using FreeSurfer\footnote{\url{http://surfer.nmr.mgh.harvard.edu/}}, with both volume- and surface-based versions of the beta weights being created for further analysis and interpretation. 
We masked preprocessed fMRI signals using the provided NSDGeneral ROI (Region-of-Interest) mask in 1.8 mm resolution. The ROI consists of 15,724, 14,278, 13,039, and 12,682 voxels for the 4 subjects respectively, and includes many visual areas from the early visual cortex to higher visual areas.
To handle the different voxel numbers of different subjects, all fMRI data are first padded to the maximum length in a wrap-around manner. Additionally, training fMRI is normalized to have zero mean and unit standard deviation.

\section{More Reconstruction Results}\label{sec:reconstruction}
\begin{figure*}[t]
    \centering    
        \vspace{-10pt}
\includegraphics[width=0.90\linewidth]{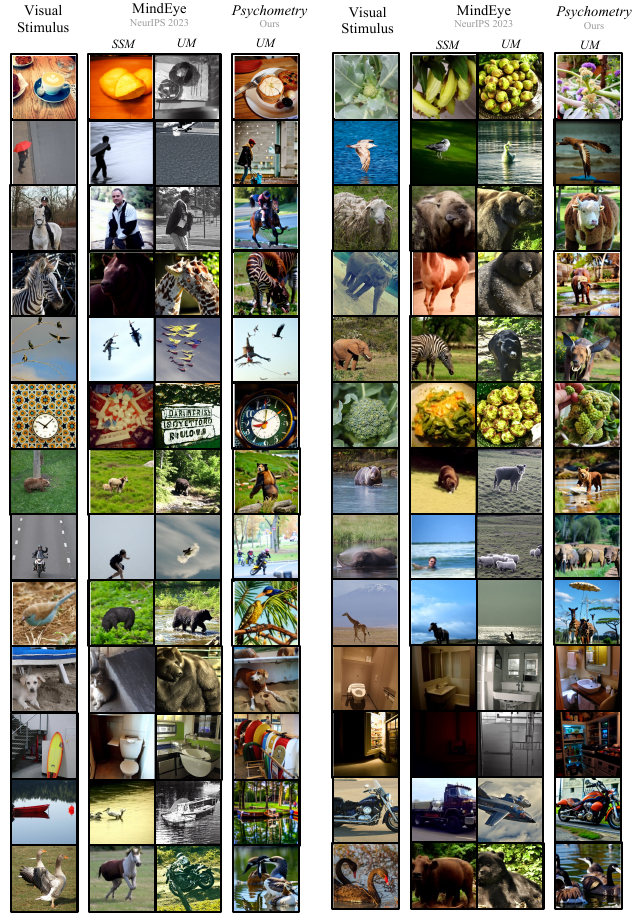}
    \caption{Additional visual comparison results on NSD 
    \texttt{test}.}
    \label{fig:vis_supp}
    \vspace{-10pt}
\end{figure*}

We provide more visual results that compare \textit{Psychometry} to the state-of-the-art  (MindEye~\cite{scotti2023reconstructing}) in Figure~\ref{fig:vis_supp}. As can be seen, \textit{Psychometry} is able to reconstruct high-quality and realistic images when utilizing a unified model (\textit{UM}) trained on the amalgamated fMRI data from different subjects.

\section{More Implementation Details}\label{sec:implementation}
The framework of the fMRI representation learning network used in our \textit{Psychometry} model is shown in Figure~\ref{fig:framework}. Before training the network, we employ a RoI embedding layer~\cite{2023lea,chen2023seeing} to separately encode fMRI signals into patches from different RoI regions. The encoded features are utilized as input to the transformer blocks.
We present the PyTorch implementation of both our \textit{Omni MoE} layer and the \textit{Ecphory} inference strategy in Algorithm~\ref{alg:soft_moe_python} and Algorithm~\ref{alg:ecphory_python} respectively. During training, we train the model for 200 epochs with a batch size of 64. We update the parameters using AdamW with $\beta_1\!=\!0.9$, $\beta_2\!=\!0.9999$, $\epsilon\!=\!10^{-8}$, and $lr\!=\!5\!\times\!10^{-4}$. During inference, $\alpha$ as the mix-up weight to enhance the output embedding is set as $0.5$.
When conducting fMRI-to-image generation, the forward and reverse diffusion blocks in the Versatile Diffusion model~\cite{xu2023versatile} are all pretrained and frozen.
To ensure reproducibility and foster future research, our full implementation will be released after acceptance.

\subsection{Omni MoE Algorithm}\label{sec:omnimoe}

\begin{lstlisting}[language=Python, caption={Pytorch implementation of our Omni MoE layer}, label=alg:soft_moe_python, escapechar=|]
def omni_moe(inputs, mix_experts, ssparams, subj_name):
  # Obtain the subject-specific parameter
  subj_param = ssparams[subj_name]
  # Compute subject-specific features
  subj_feat  = torch.einsum("bmc,ce->bme", inputs, subj_param)
  # Compute splitting weights, Eq. 2
  sweights   = torch.nn.softmax(subj_feat, dim=1)
  # Obtain the splitted features, Eq.3 
  m_feat     = torch.einsum("bmc,bme->bec", inputs, sweights)
  q_feat     = torch.stack([f_e(m_feat[:, i, :]) for i, f_e in enumerate(mix_experts)], dim=1)
  # Compute lumping weights, Eq. 4
  lweights   = torch.nn.softmax(subj_feat, dim=2)
  # Compute output tokens, Eq. 5
  outputs    = torch.einsum("bec,bme->bmc", q_feat, lweights)
  return outputs
\end{lstlisting}

\subsection{Ecphory Algorithm}\label{sec:ecphory}

\begin{lstlisting}[language=Python, caption={Pytorch implementation of our Ecphory strategy}, label=alg:ecphory_python, escapechar=|]
def ecphory(predictions, memories, K, alpha):
  # Normalize predictions and memories
  prediction_norm = F.normalize(predictions[:,0],dim=-1)
  memory_norm = F.normalize(memories[:,0],dim=-1)
  # Compute the similarities
  similarity    = prediction_norm @ memory_norm.T
  # Select memories with top K similarity scores
  topK_indexes  = torch.topk(similarity.flatten(), K).indices
  # In practice, we set K=1
  en_prediction = alpha * predictions + (1-alpha) * memories[topK_indexes]
  return en_prediction
\end{lstlisting}

\section{Social Impacts}\label{sec:broaderimpact}
This paper introduces \textit{Psychometry}, a unified model for reconstructing images from human brain activity via fMRI data. The model is designed to be omnifit, capable of handling data from various individuals, and represents a significant advancement in the field of Brain-Computer Interface.
It also has the potential to greatly enhance our understanding of brain function and could lead to breakthroughs in medical diagnostics and treatment, particularly in neurological disorders. Its ability to capture both inter-subject commonality and individual specificity can contribute to personalized medicine approaches.
Ultimately, technology capable of interpreting human brain activity is expected to revolutionize how we interact with technology, leading to new interfaces that directly connect with human thought.

\section{Limitations}\label{sec:limitation}
Currently, \textit{Psychometry} is specifically designed for image reconstruction via fMRI data. It is interesting to extend our method to handle more complicated human brain activity signals, for example, magnetoencephalography (MEG) and electroencephalography (EEG) signals. 
Moreover, while \textit{Psychometry} offers substantial benefits, it is crucial to ensure that appropriate safeguards are in place to protect individual privacy, especially when amalgamating sensitive fMRI data from different subjects.
We leave these as a part of our future work.

%% file: main.bbl
\begin{thebibliography}{71}
\providecommand{\natexlab}[1]{#1}
\providecommand{\url}[1]{\texttt{#1}}
\expandafter\ifx\csname urlstyle\endcsname\relax
  \providecommand{\doi}[1]{doi: #1}\else
  \providecommand{\doi}{doi: \begingroup \urlstyle{rm}\Url}\fi

\bibitem[Ahmed et~al.(2016)Ahmed, Baig, and Torresani]{ahmed2016network}
Karim Ahmed, Mohammad~Haris Baig, and Lorenzo Torresani.
\newblock Network of experts for large-scale image categorization.
\newblock In \emph{ECCV}, 2016.

\bibitem[Allen et~al.(2022)Allen, St-Yves, Wu, Breedlove, Prince, Dowdle, Nau, Caron, Pestilli, Charest, et~al.]{allen2022massive}
Emily~J Allen, Ghislain St-Yves, Yihan Wu, Jesse~L Breedlove, Jacob~S Prince, Logan~T Dowdle, Matthias Nau, Brad Caron, Franco Pestilli, Ian Charest, et~al.
\newblock A massive 7t fmri dataset to bridge cognitive neuroscience and artificial intelligence.
\newblock \emph{Nature Neuroscience}, 25\penalty0 (1):\penalty0 116--126, 2022.

\bibitem[Amunts and Zilles(2015)]{amunts2015architectonic}
Katrin Amunts and Karl Zilles.
\newblock Architectonic mapping of the human brain beyond brodmann.
\newblock \emph{Neuron}, 88\penalty0 (6):\penalty0 1086--1107, 2015.

\bibitem[Beliy et~al.(2019)Beliy, Gaziv, Hoogi, Strappini, Golan, and Irani]{beliy2019voxels}
Roman Beliy, Guy Gaziv, Assaf Hoogi, Francesca Strappini, Tal Golan, and Michal Irani.
\newblock From voxels to pixels and back: Self-supervision in natural-image reconstruction from fmri.
\newblock In \emph{NeurIPS}, 2019.

\bibitem[Bullmore and Sporns(2009)]{bullmore2009complex}
Ed Bullmore and Olaf Sporns.
\newblock Complex brain networks: graph theoretical analysis of structural and functional systems.
\newblock \emph{Nature Reviews Neuroscience}, 10\penalty0 (3):\penalty0 186--198, 2009.

\bibitem[Caron et~al.(2020)Caron, Misra, Mairal, Goyal, Bojanowski, and Joulin]{caron2020unsupervised}
Mathilde Caron, Ishan Misra, Julien Mairal, Priya Goyal, Piotr Bojanowski, and Armand Joulin.
\newblock Unsupervised learning of visual features by contrasting cluster assignments.
\newblock In \emph{NeurIPS}, 2020.

\bibitem[Chang et~al.(2019)Chang, Pyles, Marcus, Gupta, Tarr, and Aminoff]{chang2019bold5000}
Nadine Chang, John~A Pyles, Austin Marcus, Abhinav Gupta, Michael~J Tarr, and Elissa~M Aminoff.
\newblock Bold5000, a public fmri dataset while viewing 5000 visual images.
\newblock \emph{Scientific data}, 6\penalty0 (1):\penalty0 49, 2019.

\bibitem[Chen et~al.(1999)Chen, Xu, and Chi]{chen1999improved}
Ke Chen, Lei Xu, and Huisheng Chi.
\newblock Improved learning algorithms for mixture of experts in multiclass classification.
\newblock \emph{Neural Networks}, 12\penalty0 (9):\penalty0 1229--1252, 1999.

\bibitem[Chen et~al.(2023{\natexlab{a}})Chen, Qing, Xiang, Yue, and Zhou]{chen2023seeing}
Zijiao Chen, Jiaxin Qing, Tiange Xiang, Wan~Lin Yue, and Juan~Helen Zhou.
\newblock Seeing beyond the brain: Conditional diffusion model with sparse masked modeling for vision decoding.
\newblock In \emph{CVPR}, 2023{\natexlab{a}}.

\bibitem[Chen et~al.(2023{\natexlab{b}})Chen, Shen, Ding, Chen, Zhao, Learned-Miller, and Gan]{chen2023mod}
Zitian Chen, Yikang Shen, Mingyu Ding, Zhenfang Chen, Hengshuang Zhao, Erik~G Learned-Miller, and Chuang Gan.
\newblock Mod-squad: Designing mixtures of experts as modular multi-task learners.
\newblock In \emph{CVPR}, 2023{\natexlab{b}}.

\bibitem[Cox and Savoy(2003)]{cox2003functional}
David~D Cox and Robert~L Savoy.
\newblock Functional magnetic resonance imaging (fmri)``brain reading'': detecting and classifying distributed patterns of fmri activity in human visual cortex.
\newblock \emph{Neuroimage}, 19\penalty0 (2):\penalty0 261--270, 2003.

\bibitem[Dosovitskiy et~al.(2021)Dosovitskiy, Beyer, Kolesnikov, Weissenborn, Zhai, Unterthiner, Dehghani, Minderer, Heigold, Gelly, et~al.]{dosovitskiy2020image}
Alexey Dosovitskiy, Lucas Beyer, Alexander Kolesnikov, Dirk Weissenborn, Xiaohua Zhai, Thomas Unterthiner, Mostafa Dehghani, Matthias Minderer, Georg Heigold, Sylvain Gelly, et~al.
\newblock An image is worth 16x16 words: Transformers for image recognition at scale.
\newblock In \emph{ICLR}, 2021.

\bibitem[Enzweiler and Gavrila(2011)]{enzweiler2011multilevel}
Markus Enzweiler and Dariu~M Gavrila.
\newblock A multilevel mixture-of-experts framework for pedestrian classification.
\newblock \emph{IEEE Transactions on Image Processing}, 20\penalty0 (10):\penalty0 2967--2979, 2011.

\bibitem[Fan et~al.(2022)Fan, Sarkar, Jiang, Chen, Zou, Cheng, Hao, Wang, et~al.]{fan2022m3vit}
Zhiwen Fan, Rishov Sarkar, Ziyu Jiang, Tianlong Chen, Kai Zou, Yu Cheng, Cong Hao, Zhangyang Wang, et~al.
\newblock M$^3$vit: Mixture-of-experts vision transformer for efficient multi-task learning with model-accelerator co-design.
\newblock In \emph{NeurIPS}, pages 28441--28457, 2022.

\bibitem[Fedorenko et~al.(2010)Fedorenko, Hsieh, Nieto-Casta{\~n}{\'o}n, Whitfield-Gabrieli, and Kanwisher]{fedorenko2010new}
Evelina Fedorenko, Po-Jang Hsieh, Alfonso Nieto-Casta{\~n}{\'o}n, Susan Whitfield-Gabrieli, and Nancy Kanwisher.
\newblock New method for fmri investigations of language: defining rois functionally in individual subjects.
\newblock \emph{Journal of Neurophysiology}, 104\penalty0 (2):\penalty0 1177--1194, 2010.

\bibitem[Fedus et~al.(2022)Fedus, Zoph, and Shazeer]{fedus2022switch}
William Fedus, Barret Zoph, and Noam Shazeer.
\newblock Switch transformers: Scaling to trillion parameter models with simple and efficient sparsity.
\newblock \emph{The Journal of Machine Learning Research}, 23\penalty0 (1):\penalty0 5232--5270, 2022.

\bibitem[Frankland et~al.(2019)Frankland, Josselyn, and K{\"o}hler]{frankland2019neurobiological}
Paul~W Frankland, Sheena~A Josselyn, and Stefan K{\"o}hler.
\newblock The neurobiological foundation of memory retrieval.
\newblock \emph{Nature Neuroscience}, 22\penalty0 (10):\penalty0 1576--1585, 2019.

\bibitem[Frost and Goebel(2012)]{frost2012measuring}
Martin~A Frost and Rainer Goebel.
\newblock Measuring structural--functional correspondence: spatial variability of specialised brain regions after macro-anatomical alignment.
\newblock \emph{Neuroimage}, 59\penalty0 (2):\penalty0 1369--1381, 2012.

\bibitem[Fujiwara et~al.(2013)Fujiwara, Miyawaki, and Kamitani]{fujiwara2013modular}
Yusuke Fujiwara, Yoichi Miyawaki, and Yukiyasu Kamitani.
\newblock Modular encoding and decoding models derived from bayesian canonical correlation analysis.
\newblock \emph{Neural Computation}, 25\penalty0 (4):\penalty0 979--1005, 2013.

\bibitem[Gordon et~al.(2017)Gordon, Laumann, Adeyemo, Gilmore, Nelson, Dosenbach, and Petersen]{gordon2017individual}
Evan~M Gordon, Timothy~O Laumann, Babatunde Adeyemo, Adrian~W Gilmore, Steven~M Nelson, Nico~UF Dosenbach, and Steven~E Petersen.
\newblock Individual-specific features of brain systems identified with resting state functional correlations.
\newblock \emph{Neuroimage}, 146:\penalty0 918--939, 2017.

\bibitem[Gu et~al.(2023)Gu, Jamison, Kuceyeski, and Sabuncu]{gu2022decoding}
Zijin Gu, Keith Jamison, Amy Kuceyeski, and Mert Sabuncu.
\newblock Decoding natural image stimuli from fmri data with a surface-based convolutional network.
\newblock In \emph{MIDL}, 2023.

\bibitem[Gutta et~al.(2000)Gutta, Huang, Jonathon, and Wechsler]{gutta2000mixture}
Srinivas Gutta, Jeffrey~RJ Huang, P Jonathon, and Harry Wechsler.
\newblock Mixture of experts for classification of gender, ethnic origin, and pose of human faces.
\newblock \emph{IEEE Transactions on Neural Networks}, 11\penalty0 (4):\penalty0 948--960, 2000.

\bibitem[Hazimeh et~al.(2021)Hazimeh, Zhao, Chowdhery, Sathiamoorthy, Chen, Mazumder, Hong, and Chi]{hazimeh2021dselect}
Hussein Hazimeh, Zhe Zhao, Aakanksha Chowdhery, Maheswaran Sathiamoorthy, Yihua Chen, Rahul Mazumder, Lichan Hong, and Ed Chi.
\newblock Dselect-k: Differentiable selection in the mixture of experts with applications to multi-task learning.
\newblock In \emph{NeurIPS}, 2021.

\bibitem[Ho et~al.(2020)Ho, Jain, and Abbeel]{ho2020denoising}
Jonathan Ho, Ajay Jain, and Pieter Abbeel.
\newblock Denoising diffusion probabilistic models.
\newblock In \emph{NeurIPS}, pages 6840--6851, 2020.

\bibitem[Horikawa and Kamitani(2017)]{horikawa2017generic}
Tomoyasu Horikawa and Yukiyasu Kamitani.
\newblock Generic decoding of seen and imagined objects using hierarchical visual features.
\newblock \emph{Nature Communications}, 8\penalty0 (1):\penalty0 15037, 2017.

\bibitem[Jacobs et~al.(1991)Jacobs, Jordan, Nowlan, and Hinton]{jacobs1991adaptive}
Robert~A Jacobs, Michael~I Jordan, Steven~J Nowlan, and Geoffrey~E Hinton.
\newblock Adaptive mixtures of local experts.
\newblock \emph{Neural Computation}, 3\penalty0 (1):\penalty0 79--87, 1991.

\bibitem[Kamitani and Tong(2005)]{kamitani2005decoding}
Yukiyasu Kamitani and Frank Tong.
\newblock Decoding the visual and subjective contents of the human brain.
\newblock \emph{Nature Neuroscience}, 8\penalty0 (5):\penalty0 679--685, 2005.

\bibitem[Kay et~al.(2008)Kay, Naselaris, Prenger, and Gallant]{kay2008identifying}
Kendrick~N Kay, Thomas Naselaris, Ryan~J Prenger, and Jack~L Gallant.
\newblock Identifying natural images from human brain activity.
\newblock \emph{Nature}, 452\penalty0 (7185):\penalty0 352--355, 2008.

\bibitem[Kwong et~al.(1992)Kwong, Belliveau, Chesler, Goldberg, Weisskoff, Poncelet, Kennedy, Hoppel, Cohen, and Turner]{kwong1992dynamic}
Kenneth~K Kwong, John~W Belliveau, David~A Chesler, Inna~E Goldberg, Robert~M Weisskoff, Brigitte~P Poncelet, David~N Kennedy, Bernice~E Hoppel, Mark~S Cohen, and Robert Turner.
\newblock Dynamic magnetic resonance imaging of human brain activity during primary sensory stimulation.
\newblock \emph{Proceedings of the National Academy of Sciences}, 89\penalty0 (12):\penalty0 5675--5679, 1992.

\bibitem[LeCun et~al.(2015)LeCun, Bengio, and Hinton]{lecun2015deep}
Yann LeCun, Yoshua Bengio, and Geoffrey Hinton.
\newblock Deep learning.
\newblock \emph{Nature}, 521\penalty0 (7553):\penalty0 436--444, 2015.

\bibitem[Li et~al.(2023)Li, Quan, Zhu, and Yang]{li2023efficient}
Yaowei Li, Ruijie Quan, Linchao Zhu, and Yi Yang.
\newblock Efficient multimodal fusion via interactive prompting.
\newblock In \emph{CVPR}, 2023.

\bibitem[Lin et~al.(2022)Lin, Sprague, and Singh]{lin2022mind}
Sikun Lin, Thomas Sprague, and Ambuj~K Singh.
\newblock Mind reader: Reconstructing complex images from brain activities.
\newblock In \emph{NeurIPS}, 2022.

\bibitem[Lin et~al.(2014)Lin, Maire, Belongie, Hays, Perona, Ramanan, Doll{\'a}r, and Zitnick]{lin2014microsoft}
Tsung-Yi Lin, Michael Maire, Serge Belongie, James Hays, Pietro Perona, Deva Ramanan, Piotr Doll{\'a}r, and C~Lawrence Zitnick.
\newblock Microsoft coco: Common objects in context.
\newblock In \emph{ECCV}, 2014.

\bibitem[Liu et~al.(2023{\natexlab{a}})Liu, Wang, Wang, and Yang]{liu2023bird}
Rui Liu, Xiaohan Wang, Wenguan Wang, and Yi Yang.
\newblock Bird's-eye-view scene graph for vision-language navigation.
\newblock In \emph{ICCV}, 2023{\natexlab{a}}.

\bibitem[Liu et~al.(2023{\natexlab{b}})Liu, Ma, Zhou, Zhu, and Zheng]{liu2023brainclip}
Yulong Liu, Yongqiang Ma, Wei Zhou, Guibo Zhu, and Nanning Zheng.
\newblock Brainclip: Bridging brain and visual-linguistic representation via clip for generic natural visual stimulus decoding from fmri.
\newblock \emph{arXiv preprint arXiv:2302.12971}, 2023{\natexlab{b}}.

\bibitem[Lu et~al.(2023)Lu, Du, Zhou, Wang, and He]{lu2023minddiffuser}
Yizhuo Lu, Changde Du, Qiongyi Zhou, Dianpeng Wang, and Huiguang He.
\newblock Minddiffuser: Controlled image reconstruction from human brain activity with semantic and structural diffusion.
\newblock In \emph{ACM MM}, 2023.

\bibitem[Lu et~al.(2024)Lu, Quan, Zhu, and Yang]{lu2024zero}
Yu Lu, Ruijie Quan, Linchao Zhu, and Yi Yang.
\newblock Zero-shot video grounding with pseudo query lookup and verification.
\newblock \emph{IEEE Transactions on Image Processing}, 33:\penalty0 1643--1654, 2024.

\bibitem[Ma et~al.(2023)Ma, Jin, Wang, Huang, Zhu, Feng, and Yang]{ma2023temporal}
Fan Ma, Xiaojie Jin, Heng Wang, Jingjia Huang, Linchao Zhu, Jiashi Feng, and Yi Yang.
\newblock Temporal perceiving video-language pre-training.
\newblock In \emph{AAAI}, 2023.

\bibitem[Ma et~al.(2018)Ma, Zhao, Yi, Chen, Hong, and Chi]{ma2018modeling}
Jiaqi Ma, Zhe Zhao, Xinyang Yi, Jilin Chen, Lichan Hong, and Ed~H Chi.
\newblock Modeling task relationships in multi-task learning with multi-gate mixture-of-experts.
\newblock In \emph{ACM SIGKDD}, pages 1930--1939, 2018.

\bibitem[Mozafari et~al.(2020)Mozafari, Reddy, and VanRullen]{mozafari2020reconstructing}
Milad Mozafari, Leila Reddy, and Rufin VanRullen.
\newblock Reconstructing natural scenes from fmri patterns using bigbigan.
\newblock In \emph{IJCNN}, pages 1--8, 2020.

\bibitem[Naselaris et~al.(2009)Naselaris, Prenger, Kay, Oliver, and Gallant]{naselaris2009bayesian}
Thomas Naselaris, Ryan~J Prenger, Kendrick~N Kay, Michael Oliver, and Jack~L Gallant.
\newblock Bayesian reconstruction of natural images from human brain activity.
\newblock \emph{Neuron}, 63\penalty0 (6):\penalty0 902--915, 2009.

\bibitem[Nieto-Casta{\~n}{\'o}n and Fedorenko(2012)]{nieto2012subject}
Alfonso Nieto-Casta{\~n}{\'o}n and Evelina Fedorenko.
\newblock Subject-specific functional localizers increase sensitivity and functional resolution of multi-subject analyses.
\newblock \emph{Neuroimage}, 63\penalty0 (3):\penalty0 1646--1669, 2012.

\bibitem[Ozcelik and VanRullen(2023)]{ozcelik2023brain}
Furkan Ozcelik and Rufin VanRullen.
\newblock Brain-diffuser: Natural scene reconstruction from fmri signals using generative latent diffusion.
\newblock \emph{arXiv preprint arXiv:2303.05334}, 2023.

\bibitem[Ozcelik et~al.(2022)Ozcelik, Choksi, Mozafari, Reddy, and VanRullen]{ozcelik2022reconstruction}
Furkan Ozcelik, Bhavin Choksi, Milad Mozafari, Leila Reddy, and Rufin VanRullen.
\newblock Reconstruction of perceived images from fmri patterns and semantic brain exploration using instance-conditioned gans.
\newblock In \emph{IJCNN}, pages 1--8, 2022.

\bibitem[Parthasarathy et~al.(2017)Parthasarathy, Batty, Falcon, Rutten, Rajpal, Chichilnisky, and Paninski]{parthasarathy2017neural}
Nikhil Parthasarathy, Eleanor Batty, William Falcon, Thomas Rutten, Mohit Rajpal, EJ Chichilnisky, and Liam Paninski.
\newblock Neural networks for efficient bayesian decoding of natural images from retinal neurons.
\newblock In \emph{NeurIPS}, 2017.

\bibitem[Qian et~al.(2023)Qian, Wang, Fu, Sun, Feng, and Xue]{2023lea}
Xuelin Qian, Yikai Wang, Yanwei Fu, Xinwei Sun, Jianfeng Feng, and Xiangyang Xue.
\newblock Joint fmri decoding and encoding with latent embedding alignment.
\newblock \emph{arXiv preprint arXiv:2303.14730}, 2023.

\bibitem[Radford et~al.(2021)Radford, Kim, Hallacy, Ramesh, Goh, Agarwal, Sastry, Askell, Mishkin, Clark, et~al.]{radford2021learning}
Alec Radford, Jong~Wook Kim, Chris Hallacy, Aditya Ramesh, Gabriel Goh, Sandhini Agarwal, Girish Sastry, Amanda Askell, Pamela Mishkin, Jack Clark, et~al.
\newblock Learning transferable visual models from natural language supervision.
\newblock In \emph{ICML}, 2021.

\bibitem[Rakhimberdina et~al.(2021)Rakhimberdina, Jodelet, Liu, and Murata]{rakhimberdina2021natural}
Zarina Rakhimberdina, Quentin Jodelet, Xin Liu, and Tsuyoshi Murata.
\newblock Natural image reconstruction from fmri using deep learning: A survey.
\newblock \emph{Frontiers in Neuroscience}, 15:\penalty0 795488, 2021.

\bibitem[Riquelme et~al.(2021)Riquelme, Puigcerver, Mustafa, Neumann, Jenatton, Susano~Pinto, Keysers, and Houlsby]{riquelme2021scaling}
Carlos Riquelme, Joan Puigcerver, Basil Mustafa, Maxim Neumann, Rodolphe Jenatton, Andr{\'e} Susano~Pinto, Daniel Keysers, and Neil Houlsby.
\newblock Scaling vision with sparse mixture of experts.
\newblock In \emph{NeurIPS}, 2021.

\bibitem[Rombach et~al.(2022)Rombach, Blattmann, Lorenz, Esser, and Ommer]{rombach2022high}
Robin Rombach, Andreas Blattmann, Dominik Lorenz, Patrick Esser, and Bj{\"o}rn Ommer.
\newblock High-resolution image synthesis with latent diffusion models.
\newblock In \emph{CVPR}, pages 10684--10695, 2022.

\bibitem[Salvo et~al.(2021)Salvo, Holubecki, and Braga]{salvo2021correspondence}
Joseph~J Salvo, Ania~M Holubecki, and Rodrigo~M Braga.
\newblock Correspondence between functional connectivity and task-related activity patterns within the individual.
\newblock \emph{Current Opinion in Behavioral Sciences}, 40:\penalty0 178--188, 2021.

\bibitem[Scotti et~al.(2023)Scotti, Banerjee, Goode, Shabalin, Nguyen, Cohen, Dempster, Verlinde, Yundler, Weisberg, et~al.]{scotti2023reconstructing}
Paul~S Scotti, Atmadeep Banerjee, Jimmie Goode, Stepan Shabalin, Alex Nguyen, Ethan Cohen, Aidan~J Dempster, Nathalie Verlinde, Elad Yundler, David Weisberg, et~al.
\newblock Reconstructing the mind's eye: fmri-to-image with contrastive learning and diffusion priors.
\newblock In \emph{NeurIPS}, 2023.

\bibitem[Shazeer et~al.(2017)Shazeer, Mirhoseini, Maziarz, Davis, Le, Hinton, and Dean]{shazeer2017outrageously}
Noam Shazeer, Azalia Mirhoseini, Krzysztof Maziarz, Andy Davis, Quoc Le, Geoffrey Hinton, and Jeff Dean.
\newblock Outrageously large neural networks: The sparsely-gated mixture-of-experts layer.
\newblock In \emph{ICLR}, 2017.

\bibitem[Shen et~al.(2019{\natexlab{a}})Shen, Dwivedi, Majima, Horikawa, and Kamitani]{shen2019end}
Guohua Shen, Kshitij Dwivedi, Kei Majima, Tomoyasu Horikawa, and Yukiyasu Kamitani.
\newblock End-to-end deep image reconstruction from human brain activity.
\newblock \emph{Frontiers in Computational Neuroscience}, 13:\penalty0 21, 2019{\natexlab{a}}.

\bibitem[Shen et~al.(2019{\natexlab{b}})Shen, Horikawa, Majima, and Kamitani]{shen2019deep}
Guohua Shen, Tomoyasu Horikawa, Kei Majima, and Yukiyasu Kamitani.
\newblock Deep image reconstruction from human brain activity.
\newblock \emph{PLoS Computational Biology}, 15\penalty0 (1):\penalty0 e1006633, 2019{\natexlab{b}}.

\bibitem[Shen et~al.(2024)Shen, Ma, Zhou, and Yang]{shen2023controllable}
Xiaolong Shen, Jianxin Ma, Chang Zhou, and Zongxin Yang.
\newblock Controllable 3d face generation with conditional style code diffusion.
\newblock In \emph{AAAI}, 2024.

\bibitem[Sohl-Dickstein et~al.(2015)Sohl-Dickstein, Weiss, Maheswaranathan, and Ganguli]{sohl2015deep}
Jascha Sohl-Dickstein, Eric Weiss, Niru Maheswaranathan, and Surya Ganguli.
\newblock Deep unsupervised learning using nonequilibrium thermodynamics.
\newblock In \emph{ICML}, pages 2256--2265, 2015.

\bibitem[Song and Ermon(2019)]{song2019generative}
Yang Song and Stefano Ermon.
\newblock Generative modeling by estimating gradients of the data distribution.
\newblock \emph{NeurIPS}, 32, 2019.

\bibitem[Song et~al.(2020)Song, Sohl-Dickstein, Kingma, Kumar, Ermon, and Poole]{song2020score}
Yang Song, Jascha Sohl-Dickstein, Diederik~P Kingma, Abhishek Kumar, Stefano Ermon, and Ben Poole.
\newblock Score-based generative modeling through stochastic differential equations.
\newblock \emph{ICLR}, 2020.

\bibitem[Steinvorth et~al.(2006)Steinvorth, Corkin, and Halgren]{steinvorth2006ecphory}
Sarah Steinvorth, Suzanne Corkin, and Eric Halgren.
\newblock Ecphory of autobiographical memories: an fmri study of recent and remote memory retrieval.
\newblock \emph{Neuroimage}, 30\penalty0 (1):\penalty0 285--298, 2006.

\bibitem[Suo et~al.(2024)Suo, Ma, Zhu, and Yang]{suo2024knowledge}
Yucheng Suo, Fan Ma, Linchao Zhu, and Yi Yang.
\newblock Knowledge-enhanced dual-stream zero-shot composed image retrieval.
\newblock In \emph{CVPR}, 2024.

\bibitem[Takagi and Nishimoto(2023)]{takagi2023high}
Yu Takagi and Shinji Nishimoto.
\newblock High-resolution image reconstruction with latent diffusion models from human brain activity.
\newblock In \emph{CVPR}, 2023.

\bibitem[Tan and Le(2019)]{tan2019efficientnet}
Mingxing Tan and Quoc Le.
\newblock Efficientnet: Rethinking model scaling for convolutional neural networks.
\newblock In \emph{ICML}, 2019.

\bibitem[Taylor and Nitschke(2018)]{taylor2018improving}
Luke Taylor and Geoff Nitschke.
\newblock Improving deep learning with generic data augmentation.
\newblock In \emph{2018 IEEE symposium series on computational intelligence (SSCI)}, pages 1542--1547, 2018.

\bibitem[Tulving(1983)]{tulving1983ecphoric}
Endel Tulving.
\newblock Ecphoric processes in episodic memory.
\newblock \emph{Philosophical Transactions of the Royal Society of London. B, Biological Sciences}, 302\penalty0 (1110):\penalty0 361--371, 1983.

\bibitem[Van~Essen et~al.(2013)Van~Essen, Smith, Barch, Behrens, Yacoub, Ugurbil, Consortium, et~al.]{van2013wu}
David~C Van~Essen, Stephen~M Smith, Deanna~M Barch, Timothy~EJ Behrens, Essa Yacoub, Kamil Ugurbil, Wu-Minn~HCP Consortium, et~al.
\newblock The wu-minn human connectome project: an overview.
\newblock \emph{Neuroimage}, 80:\penalty0 62--79, 2013.

\bibitem[Xu et~al.(2023)Xu, Wang, Zhang, Wang, and Shi]{xu2023versatile}
Xingqian Xu, Zhangyang Wang, Gong Zhang, Kai Wang, and Humphrey Shi.
\newblock Versatile diffusion: Text, images and variations all in one diffusion model.
\newblock In \emph{CVPR}, 2023.

\bibitem[Yang et~al.(2024)Yang, Chen, Li, Wang, and Yang]{yang2024doraemongpt}
Zongxin Yang, Guikun Chen, Xiaodi Li, Wenguan Wang, and Yi Yang.
\newblock Doraemongpt: Toward understanding dynamic scenes with large language models.
\newblock \emph{arXiv preprint arXiv:2401.08392}, 2024.

\bibitem[Ye and Xu(2023)]{ye2023taskexpert}
Hanrong Ye and Dan Xu.
\newblock Taskexpert: Dynamically assembling multi-task representations with memorial mixture-of-experts.
\newblock In \emph{ICCV}, 2023.

\bibitem[Yosinski et~al.(2014)Yosinski, Clune, Bengio, and Lipson]{yosinski2014transferable}
Jason Yosinski, Jeff Clune, Yoshua Bengio, and Hod Lipson.
\newblock How transferable are features in deep neural networks?
\newblock In \emph{NeurIPS}, 2014.

\bibitem[Zhou et~al.(2024)Zhou, Li, Ma, Zhang, and Yang]{zhou2024migc}
Dewei Zhou, You Li, Fan Ma, Xiaoting Zhang, and Yi Yang.
\newblock Migc: Multi-instance generation controller for text-to-image synthesis.
\newblock In \emph{CVPR}, 2024.

\end{thebibliography}
